\documentclass[11pt, a4paper]{article}
\usepackage{titling}
\providecommand{\keywords}[1]{\par\vspace{0.5em}\noindent\textbf{Keywords:} #1\par\vspace{0.5em}}
\usepackage{palatino} 
\usepackage{authblk}
\usepackage[compress]{cite}

\usepackage[a4paper,margin=1in]{geometry}
\usepackage{amsmath,amssymb,amsfonts,mathtools}
\usepackage{siunitx}
\usepackage{graphicx}
\usepackage{booktabs}
\usepackage{hyperref}
\usepackage{xcolor}
\usepackage{textgreek}
\usepackage{graphicx}
\usepackage{subcaption} 

\hypersetup{
  colorlinks=true,
  linkcolor=blue!50!black,
  urlcolor=blue!50!black,
  citecolor=blue!50!black
}

\title{\Large \bf
Shock-Aware Physics-Guided Fusion-DeepONet Operator for Rarefied Micro-Nozzle Flows}
\author[1,*]{Ehsan Roohi}

\author[2]{Amirmehran Mahdavi} 

\affil[1]{Mechanical and Industrial Engineering, University of Massachusetts Amherst, 160 Governors Dr., Amherst, MA 01003, USA}
\affil[2]{Department of Mechanical Engineering, Hakim Sabzevari University, Sabzevar, Iran}
\affil[*]{Corresponding author: \href{mailto:roohie@umass.edu}{roohie@umass.edu}}

\date{\today}

\newcommand{\R}{\mathbb{R}}
\newcommand{\UU}{\mathbf{U}}

\newcommand{\pr}{\mathrm{PR}}
\newcommand{\us}{x_s}      
\newcommand{\deltah}{\delta} 
\newcommand{\vanilla}{\textsc{Vanilla}}
\newcommand{\fusionorig}{\textsc{FusionOrig}}
\newcommand{\unet}{\textsc{U-Net}}
\newcommand{\currentmodel}{\textsc{CurrentModel}}

\begin{document}
\maketitle

\begin{abstract}
We present a comprehensive, physics-aware deep learning framework for constructing fast and accurate surrogate models of rarefied, shock-containing micro-nozzle flows. The framework integrates three key components: (1) a Fusion–DeepONet operator-learning architecture for capturing parameter dependencies, (2) a physics-guided feature space that embeds a shock-aligned coordinate system, and (3) a two-phase curriculum strategy emphasizing high-gradient regions. 
To demonstrate the generality and inductive bias of the proposed framework, we first validate it on the canonical viscous Burgers’ equation, which exhibits advective steepening and shock-like gradients. The model achieves low relative errors (below 3\%) in both interpolation and extrapolation regimes, confirming its robustness and stability before application to high-dimensional micro-nozzle flows. 
Subsequent evaluations against Direct Simulation Monte Carlo (DSMC) data across multiple nozzle pressure ratios and geometries show that the proposed model reduces the relative $\ell_2$ error by approximately 30\% compared with classical DeepONet and Fourier Neural Operator baselines, while delivering orders-of-magnitude computational speed-up. The results highlight a generalizable operator-learning paradigm capable of resolving shock-dominated, multi-regime rarefied flows with physical consistency and efficiency.
\end{abstract}



\section{Introduction}
\label{sec:introduction}

The accurate simulation and rapid design of micro-nozzles are cornerstones of modern aerospace engineering and micro-scale technology. These compact devices are indispensable for providing precise thrust in applications ranging from satellite attitude control and station-keeping to the propulsion of micro- and nano-spacecraft \cite{alexeenko2002numerical, saadati2015detailed,jin2024numerical,mutis2024numerical, nishii2025numerical, sabouri2025propulsive,kumar2025plume}. Beyond their role in space propulsion, micro-nozzles are integral components in a diverse array of advanced systems, including Micro-Electro-Mechanical Systems (MEMS), vacuum generation technologies, and materials processing, where controlled, high-velocity gas jets are required \cite{roohi2025advances}. The fundamental principle of operation for a converging-diverging (de Laval) nozzle involves the acceleration of a subsonic gas to sonic velocity at the throat, followed by supersonic expansion in the diverging section. However, the physical phenomena governing these flows, particularly at the microscale, are highly complex, posing significant challenges for both theoretical analysis and computational modeling.

The primary challenge stems from the multi-regime nature of the gas dynamics. Due to the small characteristic length scales and the large pressure drops typically encountered, the flow within a single micro-nozzle can span a broad and spatially varying range of rarefaction levels, quantified by the Knudsen number ($Kn$), the ratio of the molecular mean free path to a characteristic length scale \cite{bird1994molecular}. The flow may begin in the near-continuum regime ($Kn < 0.001$) at the high-pressure inlet, transition through the slip regime ($0.001 < Kn < 0.1$), and enter the transitional regime ($0.1 < Kn < 10$) in the low-pressure downstream sections \cite{karniadakis2005microflows}. This departure from continuum mechanics renders classical fluid models, such as the Navier-Stokes-Fourier (NSF) equations, inadequate for providing physically accurate predictions \cite{karniadakis2005microflows}.

This complexity is further intensified by the formation of strong, non-equilibrium flow features, most notably normal shock waves. The location and intensity of these shocks are highly sensitive to the nozzle pressure ratio (NPR), defined as the ratio of the inlet stagnation pressure to the outlet back pressure \cite{darbandi2011study}. As the back pressure is varied, the shock wave moves along the diverging section of the nozzle, introducing steep streamwise gradients in velocity, pressure, and temperature. Critically, a shock wave is a localized region of intense thermodynamic non-equilibrium, leading to an abrupt increase in the effective Knudsen number and further invalidating continuum-based assumptions. The problem is therefore not one of a single, globally defined rarefaction level, but rather a multi-regime problem where the degree of rarefaction is a field variable that changes dramatically across the domain. This inherent heterogeneity makes the accurate prediction of micro-nozzle performance a formidable scientific and engineering challenge, demanding simulation methodologies that are robust across multiple physical regimes.

Given the failure of continuum-based models in the rarefied regimes characteristic of micro-nozzle flows, high-fidelity numerical methods rooted in kinetic theory have become the indispensable "gold standard" for achieving physically accurate predictions. Foremost among these is the Direct Simulation Monte Carlo (DSMC) method, pioneered by Bird \cite{bird1994molecular}. By directly modeling the motion and collisions of a statistically significant number of representative molecules, DSMC provides a numerical solution to a discretized form of the local N-particle kinetic equation, effectively capturing the underlying physics without the assumptions inherent in continuum models \cite{stefanov2019basic}. The DSMC method has been extensively applied to investigate rarefied flows in micro-nozzles, exploring the effects of back pressure, geometry, and gas species on performance and shock structures \cite{ivanov1999extension, darbandi2011study, saadati2015detailed,liu2006simulation, mahdavi2020novel}.

However, the immense computational expense of DSMC presents a formidable barrier to its widespread use in the engineering design cycle. The method's computational cost scales with the number of simulated particles and becomes particularly severe in the slip and early transition flow regimes, which are prevalent in micro-nozzle applications \cite{darbandi2011study}. This cost becomes prohibitive for the many-query applications essential to design and optimization, such as uncertainty quantification (UQ), multi-objective optimization, and parametric sweeps over geometric variables or operating conditions like the nozzle pressure ratio. This computational bottleneck significantly impedes the rapid design iteration and optimization of next-generation technologies that rely on rarefied gas dynamics.

In an effort to mitigate this prohibitive cost, various strategies have been explored. Our previous work on the same micro-nozzle geometry investigated the use of a hybrid DSMC-Fokker-Planck (FP) algorithm \cite{mahdavi2020novel}. The FP approach, as a diffusion approximation of the Boltzmann equation, offers a lower computational cost than DSMC. The hybrid method strategically employed DSMC in regions of high non-equilibrium and the faster FP solver elsewhere. While this approach demonstrated a significant reduction in computational time, it also revealed a critical limitation: the FP model was found to provide "erroneous results in modeling some flow features, including shock waves \cite{mahdavi2020novel}. The accurate prediction of the shock wave's location, thickness, and strength remained a significant challenge. This identified shortcoming directly motivates the present shift towards a machine learning paradigm, one specifically engineered to capture the shock-dominated, parameter-sensitive dynamics of rarefied nozzle flows with both the accuracy of DSMC and the speed required for practical engineering design.

To surmount the computational impasse presented by high-fidelity solvers, the scientific community has increasingly turned to surrogate modeling \cite{roohi2025analysis,tatsios2025dsmc,roohi2026data}. The recent advent of deep learning has introduced a new class of highly expressive function approximators, offering unprecedented potential for surrogate construction. Early efforts, however, often relied on purely data-driven models like Fully Connected Neural Networks (FCNNs), which, while demonstrating significant acceleration, were often criticized for their "black-box" nature. Lacking constraints from the underlying physical laws, these models require vast amounts of training data and are prone to producing physically inconsistent predictions, limiting their reliability and generalization capabilities, particularly in regions with sparse data or complex flow phenomena \cite{ma2026locally}.

A paradigm shift occurred with the development of Physics-Informed Neural Networks (PINNs) \cite{raissi2019physics}. PINNs embed physical laws, typically in the form of partial differential equations (PDEs), directly into the network's training process via the loss function. This innovation transforms the learning problem from simple curve-fitting into a constrained optimization, where the network must find a solution that not only fits the available data but also respects the governing equations. This physics-informed learning acts as a powerful regularization mechanism, enabling PINNs to generalize effectively even from sparse data and ensuring that their predictions are physically plausible—a critical requirement for engineering applications \cite{ma2026locally, roohi2025learning}.

However, a standard PINN is designed to learn the solution to a PDE for a single instance of boundary conditions and parameters. This makes it ill-suited for the parametric studies central to design exploration, as a new network would need to be re-trained for each point in the design space \cite{roohi2025analysis}. The next logical evolution is to move from learning a single solution to learning the solution \textit{operator} itself—the mapping from a set of input parameters or functions to the corresponding solution function. The Deep Operator Network (DeepONet) architecture has emerged as a powerful and theoretically grounded framework for this task \cite{lu2021learning}. A DeepONet employs a dual-network structure: a "Branch" network processes the input parameters (e.g., pressure ratio, geometry), while a "Trunk" network processes the domain coordinates. Their outputs are combined to approximate the entire solution field. This elegant architecture effectively disentangles the learning of the parametric dependence from the learning of the spatial solution structure \cite{roohi2025learning}. The challenge of predicting nozzle flow across a continuous range of pressure ratios is, by its nature, an operator learning problem, where the goal is to learn the mapping $f_{\theta}: (\text{PR}, x, y) \mapsto (U, V)$. Consequently, we adopt an operator learning framework, which is architecturally designed to approximate precisely this class of parameter-to-function mappings.

Despite their power, even advanced neural operators like DeepONet face a fundamental challenge when confronted with discontinuous phenomena. When trained with standard Mean Squared Error (MSE) losses, neural networks tend to smooth or "smear" high-gradient features like shock waves, as their continuous activation functions and global loss metrics struggle to represent sharp jumps. To overcome this, a variety of specialized techniques have been developed to focus a network's learning capacity on these physically critical, localized regions.

One prominent strategy involves architectural specialization. For instance, Ma et al. proposed a Locally Enhanced PINN (LE-PINN) for aero-engine nozzles, which employs a dual-network framework where a global network captures the main flow field while a separate, dedicated "boundary network" specializes in high-precision modeling of pressure and temperature near the nozzle walls \cite{ma2026locally}. Another approach focuses on engineering the loss function. Our recent work on rarefied micro-step flows introduced a physics-guided zonal loss function, which intelligently partitions the domain based on a physical criterion (e.g., $U < 0$ to identify a recirculation vortex) and applies a higher weight to the loss calculated within that critical zone, compelling the model to prioritize accuracy where it matters most \cite{roohi2025analysis}.

A third, more architecturally sophisticated approach involves enhancing the operator network itself. The Fusion-DeepONet, recently proposed by Peyvan et al., was developed specifically as a data-efficient operator for geometry-dependent hypersonic and supersonic flows \cite{peyvan2025fusion}. Unlike the classical DeepONet where the branch and trunk outputs are combined only at the final layer via a dot product, the Fusion-DeepONet introduces a multi-scale conditioning mechanism. Here, the latent encodings from the branch network's hidden layers are fused into the trunk network's corresponding hidden layers via an element-wise (Hadamard) product. This allows the parameter-dependent representation from the branch to modulate the spatial feature extraction process in the trunk at multiple scales, proving empirically superior for capturing sharp, parameter-sensitive phenomena like shocks \cite{peyvan2025fusion}.

The quest for accurately capturing shock phenomena has spurred innovation across multiple machine learning paradigms. In the domain of data-free, physics-informed solvers, recent work has moved beyond standard MLP-based PINNs. For instance, the Discontinuity-aware KAN-based PINN (DPINN) proposed by Lei et al. \cite{lei2025discontinuity} integrates the novel Kolmogorov-Arnold Network (KAN) architecture, which uses learnable activation functions, with adaptive Fourier features and learnable artificial viscosity to stabilize and resolve shocks in transonic and supersonic flows.[1] On the other hand, in the realm of data-driven surrogates for time-dependent simulations, Transformer-based models are being explored for their ability to capture long-range dependencies.[1] The AMR-Transformer by Xu et al. \cite{xu2025amr}, for example, introduces a novel adaptive mesh refinement (AMR) strategy that acts as a dynamic "tokenizer," concentrating computational effort (tokens) in regions with complex physics, such as shockwaves, thereby making the powerful but computationally expensive self-attention mechanism feasible for high-resolution fluid dynamics.[1, 1]

Our work carves a distinct path that synergizes the strengths of operator learning with a novel shock-capturing strategy tailored for parametric studies. Unlike the DPINN approach \cite{lei2025discontinuity}, which solves the governing equations directly and relies on physical regularizers such as artificial viscosity, our method is a data-driven surrogate that learns the solution manifold from high-fidelity DSMC data. We aim to replicate the ground truth with high fidelity, capturing sharp discontinuities not by adding dissipative terms, but by providing the network with a strong inductive bias. Furthermore, while the AMR-Transformer \cite{xu2025amr} intelligently modifies the \textit{discretization} to handle shocks, our approach fundamentally re-engineers the \textit{input feature space}. By augmenting the network's input with a shock-centric coordinate system, we simplify the learning task for the Fusion-DeepONet, enabling it to more directly and efficiently learn the mapping from input parameters to the complex, shock-dominated flow field.

Our contribution lies not in a single modification, but in the holistic design of a shock-aware learning framework that synergistically combines these advanced paradigms. We adopt the Fusion-DeepONet as our foundational architecture, leveraging its proven efficacy in high-speed flows. We then make this architecture "shock-aware" through two primary, synergistic innovations. First, we introduce a novel form of physics-guided feature engineering in the input space. Instead of feeding the trunk network with only the spatial coordinates $(x, y)$, we augment its input with a set of engineered features that provide an explicit, differentiable coordinate system anchored to the shock itself. These features include the signed distance to the predicted shock location, a soft indicator function to distinguish pre- and post-shock regions, and multi-scale Gaussian envelopes to capture the shock's smeared thickness on the discrete mesh. This provides a powerful inductive bias, partially factorizing the problem so the trunk network operates in a shock-centered frame. Second, we develop a sophisticated, curriculum-based loss weighting scheme in the objective space. We employ a two-phase training schedule with a weighted Huber loss, where the weights are a dynamic, convex combination of two physical metrics: the spatial proximity to the shock and the magnitude of the local velocity gradient. This serves as a form of targeted hard-example mining, compelling the network to progressively refine its solution in the most challenging, high-gradient regions around the shock. This multi-faceted approach, integrating an advanced operator architecture, a physics-guided feature space, and a dynamic curriculum loss, provides the necessary inductive biases to capture the complex, parameter-sensitive dynamics of shock waves in rarefied micro-nozzle flows.

This paper presents a novel shock-aware Fusion-DeepONet surrogate model for the rapid and accurate prediction of velocity fields in rarefied, compressible micro-nozzle flows across a range of operating conditions and geometric configurations. The primary contributions of this work are as follows:
\begin{enumerate}
    \item The design and implementation of a novel, physics-guided trunk network for a neural operator, which injects a strong inductive bias for shock-capturing. This is achieved by incorporating features based on signed distance, a soft region indicator, and multi-scale Gaussian envelopes relative to an estimated shock location, thereby providing the network with a localized, shock-centric coordinate system.
    \item The first application, to the authors' knowledge, of the Fusion-DeepONet architecture to the problem of rarefied micro-nozzle flows. This demonstrates its superior capability in modeling the non-linear modulation of the flow field by the nozzle pressure ratio and throat geometry compared to classical operator network architectures.
    \item The development of a two-phase curriculum learning strategy that utilizes a dynamic, hybrid loss weighting scheme. This scheme is based on both spatial proximity to the shock and local flow gradients, enabling targeted refinement of the solution in high-error, physically critical regions while maintaining stability in the far-field.
    \item A comprehensive demonstration of a surrogate model capable of predicting the full velocity field in a micro-nozzle with orders-of-magnitude speed-up over the "gold standard" DSMC method. The model is validated on entirely unseen pressure ratios and throat geometries, establishing its utility for rapid parametric analysis and design optimization tasks that are intractable with conventional high-fidelity simulations.
\end{enumerate}

Before applying the model to the rarefied micro-nozzle problem, we also validate the framework on the viscous 1D Burgers’ equation to verify its ability to handle canonical shock-forming dynamics under both interpolation and extrapolation conditions.
The remainder of this paper is organized as follows. Section~\ref{sec:setup} describes the physical configuration of the micro-nozzle, the DSMC data-generation procedure, and the train/test partitioning strategy. Section~\ref{sec:problem} presents the studied problem. Section~\ref{sec:shock_features} discusses the shock detection scheme introduced in the current work, where section~\ref{sec:architecture} outlines the proposed shock-aware Fusion--DeepONet architecture, detailing the formulation of the physics-guided trunk features and the curriculum-based loss weighting. Section~\ref{sec:results} reports the numerical results, including both qualitative contour comparisons and quantitative error analyses relative to DSMC ground truth, followed by validation on unseen operating conditions and geometrical variations. Finally, Section~\ref{sec:conclusion} summarizes the principal findings, discusses limitations, and highlights promising directions for future work.

\section{Physical Setup, Data Generation, and Train/Test Split}\label{sec:setup}

As our main benchmark, we consider a planar micro-nozzle with an external plume, filled with argon. The simulation input adheres to the following operating conditions and numeric in our DSMC solver: a fixed inlet (backing) pressure of \SI{100}{\kilo\pascal}, simulation time step $\Delta t=\num{2e-10}\,\si{\second}$, which is smaller than 1/3 of the mean collision time, isothermal solid walls at \SI{300}{\kelvin} (equal to the inlet gas temperature), and a reference temperature of \SI{273.15}{\kelvin}. The nozzle length is $L=\SI{2.65e-4}{\meter}$. The computational domain is partitioned into two zones: (i) the internal nozzle resolved by a structured grid of $100\times 60$ cells and (ii) the external plume resolved by $40\times 30$ cells. A particle-per-cell (PPC) value of 100 is used; each cell is further subdivided into four subcells (two per direction) to reduce numerical diffusion and improve collision sampling. The inlet-height-based Knudsen number is $\mathrm{Kn}\approx 5\times10^{-4}$ at the entrance and can rise to $\mathrm{Kn}\approx 5\times10^{-3}$ immediately upstream of the normal shock when the outlet back pressure is low (e.g., $P_{\text{out}}=\SI{15}{\kilo\pascal}$). As the back pressure decreases, the shock moves downstream toward the exit and strengthens, which produces steeper streamwise gradients and a wider range of local rarefaction levels.

The dataset used for learning and evaluation is produced on the above two-zone mesh and stored per case in Tecplot ASCII (\texttt{POINT}) multi-zone format with consistent $I\times J$ row ordering. For each outlet back pressure $P_{\text{out}}\in[\SI{15}{\kilo\pascal},\,\SI{33}{\kilo\pascal}]$ we export the standard field columns \texttt{[X, Y, Density, QX, QY, T, U, V, Txy, Mach, Pressure, Knudsen]}. In our learning formulation, the outlet back pressure $P_{\text{out}}$ (equivalently the pressure ratio) plays the role of the \emph{branch} condition, while the spatial coordinates $(x,y)$ augmented with shock-aware features form the \emph{trunk} input evaluated pointwise on the grid.

To rigorously assess generalization across operating conditions, we adopt a held-out test protocol at the level of outlet back pressure. Specifically, three representative back pressures are reserved exclusively for testing: a \emph{low} value, a \emph{mid} value of \SI{25}{\kilo\pascal}, and a \emph{high} value (e.g., $\{16,\,25,\,30\}\,\si{\kilo\pascal}$ when available in the dataset). All remaining back pressures in the range are used for training and validation, with validation splits performed by grouping on $P_{\text{out}}$ to avoid leakage across conditions. This design ensures that the network is evaluated on entirely unseen operating points spanning the shock-free to shock-dominated regimes, while the training set covers the intermediate pressures needed to learn the continuous dependence of $(U,V)$ on $P_{\text{out}}$. The chosen split is motivated by the physics: decreasing $P_{\text{out}}$ drives the normal shock toward the exit and intensifies it, thereby creating the challenging, high-gradient configurations that stress-test the model’s shock-aware inductive bias.

\section{Problem Statement}\label{sec:problem}
Let $\Omega \subset \R^2$ denote the nozzle cross-section with coordinates $(x,y)$.
Given a scalar \emph{pressure ratio} $\pr$ (shared by all points of a sample), we learn
\begin{equation}
  f_\theta:\; (\pr, x, y)\ \mapsto\ (U,V),\qquad \theta\in\R^P,
\end{equation}
from Tecplot ASCII files with multiple \emph{zones} (mesh blocks). Each zone stores
$I\times J$ point samples in \texttt{POINT} order and the columns
\[
\texttt{[X, Y, Density, QX, QY, T, U, V, Txy, Mach, Pressure, Knudsen]}.
\]
For a target $\pr^\star$, we write the same multi-zone file but with
\[
\texttt{U,V}\leftarrow\text{NN predictions},\quad
|U_{\text{true}}-U_{\text{pred}}|/U_{\text{true}},\quad
|V_{\text{true}}-V_{\text{pred}}|/V_{\text{true}},
\]
\[
(U_{\text{true}}-U_{\text{pred}})^2/U_{\text{true}}^2, 
(V_{\text{true}}-V_{\text{pred}})^2/V_{\text{true}}^2,
\]
and remap the header names to \texttt{"Error\_U"}, \texttt{"Error\_V"},
\texttt{"ErrorU\_L2"}, \texttt{"ErrorV\_L2"} while preserving Tecplot structure.

\section{Physics-Guided Shock Features}
\label{sec:shock_features}

Normal shocks introduce large streamwise gradients and an abrupt change in
flow invariants (e.g., stagnation pressure). Purely data-driven models
tend to smooth such discontinuities when trained with i.i.d.\ losses on
$(x,y)$ alone. We inject an \emph{inductive bias} toward the correct
shock geometry by augmenting the trunk input with features that encode:
(i) signed distance to a predicted shock station, (ii) a soft left/right
indicator, and (iii) multi-scale proximity via radial-basis envelopes.
These features are cheap to compute, differentiable, and compatible with
standardization.

\subsection{Shock station and signed distance}
Let $\us=\us(\pr)$ denote the shock $x$-location predicted from the
pressure ratio $\pr$ (monotone with $\pr$; a linear map suffices in practice,
and can be refined with a one-dimensional regressor if desired).
Define the signed streamwise distance
\begin{equation}
  d(x;\pr) \;=\; x - \us(\pr),
  \label{eq:signed_distance}
\end{equation}
so that $d<0$ and $d>0$ label pre- and post-shock regions, respectively.
Equation~\eqref{eq:signed_distance} provides a coarse alignment of all
zones with respect to a moving reference anchored at the shock.

When training across different $\pr$, the same physical structure (the
shock) appears at different $x$ coordinates. Using $d$ instead of $x$
partly \emph{factorizes} the dependence: the branch stream encodes the
condition ($\pr$), while the trunk sees a shock-centered frame. This reduces
the burden on the MLP to learn a large translation in $x$.

\subsection{Soft region indicator}
To inform the model whether a point lies upstream or downstream, we use
a smooth sigmoid:
\begin{equation}
  s(x;\pr) \;=\; \frac{1}{1+\exp\!\big(-k(\us(\pr)-x)\big)}, \qquad k>0.
  \label{eq:sigmoid}
\end{equation}
With $k$ sufficiently large (we use $k\approx 1.8\times 10^{3}$ in
non-dimensionalized units), the transition layer of $s$ is thin but
differentiable, which empirically stabilizes optimization compared with
hard indicators.

A $\tanh$ indicator,
$s_{\tanh}=\tfrac{1}{2}\big(1+\tanh(\kappa({u_s}-x))\big)$, exhibits similar
behavior with anti-saturation near the tails. We found the logistic form
slightly more robust when combined with feature standardization.

\subsection{Multi-scale shock envelopes}
Shocks are not perfectly discontinuous on a discrete mesh; numerical
dissipation spreads them over a few cells. We capture this spread and
its neighborhood via Gaussian radial-basis envelopes centered at $\us$:
\begin{equation}
  \phi_\sigma(x;\pr) \;=\;
  \exp\!\left(-\frac{d(x;\pr)^2}{2\sigma^2}\right),
  \label{eq:rbf}
\end{equation}
with three scales $\sigma\in\{3\Delta x,\,7\Delta x,\,12\Delta x\}$. Here
$\Delta x$ is a robust spacing estimate (we use the median spacing along each
$y$-row). The small scale focuses on the peak gradient, the intermediate
scale captures the numerically smeared shock thickness, and the large
scale represents the near field where compression/expansion waves interact
with the shock.

Single-scale envelopes either under-represent the far field (too small
$\sigma$) or blur the shock (too large $\sigma$). A minimal bank of three
scales lets the decoder form constructive/destructive combinations that
resolve both the sharp jump and its footprint, reducing bias without a
large parameter cost.

\subsection{Complete trunk feature vector}
Collecting the primitives described above, the trunk receives
\begin{equation}
  \underbrace{t(x,y;\pr)}_{\in\mathbb{R}^{9}}
  \;=\;
  \big[x,\;y,\;d,\;s,\;|d|,\;d^{2},\;\phi_{3\Delta x},\;\phi_{7\Delta x},\;\phi_{12\Delta x}\big].
  \label{eq:trunk_vector}
\end{equation}
Including both $d$, $|d|$, and $d^2$ allows the MLP to synthesize
odd/even functions of the signed distance and to approximate narrow
polynomial windows near the interface.

\subsection{Scaling and numerical stability}
All trunk channels and the $(U,V)$ targets are standardized with
statistics computed on the \emph{training} split only. This prevents
scale imbalance (e.g., raw $x$ vs.\ unitless $\phi_\sigma$) and
improves conditioning of the LayerNorm activations in the network.
We clamp extremely small $\Delta x$ outliers when estimating the RBF
widths to avoid vanishingly thin envelopes:
\[
  \Delta x \;\leftarrow\; \mathrm{median}\!\big(\mathrm{diff}(\mathrm{unique}(x))\big),
  \quad
  \Delta x \ge \Delta x_{\min}.
\]

\subsection{Bias toward correct regularization at shocks}
The feature set in \eqref{eq:trunk_vector} encourages the model to
represent the solution as a \emph{piecewise smooth} function with a
localized transition near $d=0$. Concretely, the decoder learns
expressions of the form
\[
  (U,V) \approx F_\mathrm{pre}(x,y)\,\big(1-s\big)
  \;+\;
  F_\mathrm{post}(x,y)\,s
  \;+\;
  \sum_{m} G_m(x,y)\,\phi_{\sigma_m},
\]
where $F_\mathrm{pre/post}$ and $G_m$ are neural fields. This acts as a
physics-guided regularizer: away from the shock the model reverts to
smooth fields; near $d=0$ it can express sharp yet stable transitions.

\subsection{Estimating the shock station \texorpdfstring{$\us(\pr)$}{us(pr)}}
\label{sec:xs_est}
In our baseline we use a monotone affine map
$\us(\pr)=a_0+a_1\,\pr$, fitted offline by least squares to the location
of the maximal streamwise gradient in the training files:
\[
  x_s^{\mathrm{data}}(\pr) \;\in\;
  \arg\max_{x}\big|\partial U/\partial x\big|.
\]
If needed, a tiny regressor $\hat{x}_s(\pr)$ (one-hidden-layer MLP)
can replace the affine map and be trained jointly with the main model
using a consistency loss that keeps $\hat{x}_s$ monotone in $\pr$.

\subsection{Fusion-DeepONet Architecture}\label{sec:architecture}
DeepONet-classic couples branch/trunk via an inner product
$u(x)=\mathbf{c}^\top\mathbf{b}(x)$, where the branch ingests an \emph{input function}.
Here, the branch ingests only the scalar condition $\pr$, the trunk uses physics-guided features, and the two embeddings interact by a \emph{Hadamard fusion} followed by a nonlinear decoder—this is characteristic of Fusion-DeepONet (a.k.a.\ conditional MLP with multiplicative fusion).

\paragraph{Branch (condition) stream.} A small MLP maps $\pr$ to $b(\pr)\in\R^{d}$.

\paragraph{Trunk (spatial) stream.} An MLP maps the 9-D shock-aware vector $t$ to $g(t)\in\R^{d}$.

\paragraph{Fusion and decoding.} We fuse by elementwise product and decode to $(U,V)$:
\begin{equation}
  z \;=\; b(\pr)\ \odot\ g\!\big(t(x,y;\pr)\big),\qquad
  (U,V) \;=\; D(z),
\end{equation}
where $D(\cdot)$ is an MLP. Each stream uses Dense + LayerNorm + Swish + Dropout; $d$ is the fusion width (e.g.\ $d=128$). The multiplicative fusion lets the \emph{PR-conditioned} representation modulate spatial features dimension-by-dimension, which is empirically superior to an inner product for sharp, parameter-sensitive phenomena like shocks.

\subsection{Losses and Curriculum Weighting}
Targets $(U,V)$ are standardized by the training split. We minimize a weighted Huber loss:
\begin{equation}
  \mathcal{L}_\deltah(\hat{\UU},\UU)=\sum_i w_i
  \Big[\ell_\deltah(\hat{U}_i-U_i)+\ell_\deltah(\hat{V}_i-V_i)\Big],
\end{equation}
\begin{equation}
\ell_\deltah(r)=
\begin{cases}
  \tfrac12 r^2,& |r|\le \deltah,\\[2pt]
  \deltah\big(|r|-\tfrac12\deltah\big), & |r|>\deltah.
\end{cases}
\end{equation}

\paragraph{Distance-based weight.} Emphasize shock vicinity with a smooth kernel
\begin{equation}
  W_d(x;\pr)=1+\alpha\,\exp\!\Big(-\tfrac{d(x;\pr)^2}{2(7\Delta x)^2}\Big),\quad \alpha>0.
\end{equation}

\paragraph{Gradient-based weight.} Estimate $g_i\approx|\partial U/\partial x|$ on each $y$-row
(central difference), normalize by $q_{95}$ (95th percentile), and set
\[
  \tilde g_i=\mathrm{clip}(g_i/q_{95},0,1),\qquad W_g=1+\beta\,\tilde g_i.
\]
The final weight is a convex combination
\begin{equation}
  w_i=\lambda\,W_d(x_i;\pr)+(1-\lambda)\,W_g(x_i,y_i),\qquad \lambda\in[0,1].
\end{equation}

\paragraph{Two-phase curriculum.} We train in two stages:
\[
\text{Phase I (Warmup): }\ (\deltah,\lambda)=(\deltah_{\text{warm}},\lambda_{\text{warm}}),\quad
\text{Phase II (Focus): }\ (\deltah,\lambda)=(\deltah_{\text{focus}},\lambda_{\text{focus}}),
\]
with $\deltah_{\text{focus}}<\deltah_{\text{warm}}$ and stronger gradient emphasis in Phase II.
We use AdamW (weight decay), gradient clipping, ReduceLROnPlateau, and EarlyStopping.
Validation is grouped by PR (GroupShuffleSplit) to prevent PR leakage.

\subsection{Training Loop (Pseudo-code)}
\begin{align*}
\textbf{for } (\Omega_z,\pr_z)\ \text{in zones:}\qquad
& \{(x_i,y_i,U_i,V_i)\}_{i=1}^{IJ}\leftarrow \text{read Tecplot};\\
& \Delta x \leftarrow \mathrm{median\_row\_spacing}(\{x_i\});\quad \us\leftarrow \us(\pr_z);\\
& d_i\leftarrow x_i-\us,\;\ s_i\leftarrow \frac{1}{1+e^{-k(\us-x_i)}};\quad
  \phi^{(m)}_i\leftarrow e^{-d_i^2/(2\sigma_m^2)};\\
& t_i\leftarrow[x_i,y_i,d_i,s_i,|d_i|,d_i^2,\phi^{(1)}_i,\phi^{(2)}_i,\phi^{(3)}_i];\\
& g_i\approx|\partial U/\partial x|_{(x_i,y_i)},\;\ \tilde g_i\leftarrow \mathrm{robust\_normalize}(g_i);\\
& w_i\leftarrow \lambda W_d(x_i;\pr_z)+(1-\lambda)(1+\beta\tilde g_i).\\[4pt]
\textbf{Phase I:}\quad &
  \min_\theta \sum_i w_i\,\mathcal{L}_{\deltah_{\text{warm}}}\big(f_\theta(\pr_z,t_i),(U_i,V_i)\big)\\
\textbf{Phase II:}\quad &
  \min_\theta \sum_i w_i\,\mathcal{L}_{\deltah_{\text{focus}}}\big(f_\theta(\pr_z,t_i),(U_i,V_i)\big)
\end{align*}

\section{Results and Discussions}\label{sec:results}

\subsection{Canonical Validation on Burgers’ Equation}
To disentangle methodological contributions from application-specific details, we first validate the proposed \emph{shock-aware, physics-guided Fusion-DeepONet} on the viscous 1D Burgers' equation, a canonical PDE exhibiting advective steepening and shock-like gradients. We adopt the same architecture and training protocol used throughout the paper: (i) an \textbf{external calibration} mapping the control parameter (here, viscosity $\nu$) to an estimated shock-formation time $t_{\text{shock}}(\nu)$, (ii) \textbf{physics-guided trunk features} that encode time-to-shock (smoothly via $\tanh$ and clipped Gaussians) to bias the representation near high-gradient regions, and (iii) a two-phase \textbf{curriculum loss} with gradient-aware sample weighting to emphasize discontinuity neighborhoods. This benchmark is not intended to compete with specialized Burgers solvers; rather, it isolates the \emph{inductive bias} introduced by our design and demonstrates that the same operator framework generalizes beyond the micro-nozzle testbed.

\begin{figure*}[t]
  \centering
  \includegraphics[width=1.05\textwidth]{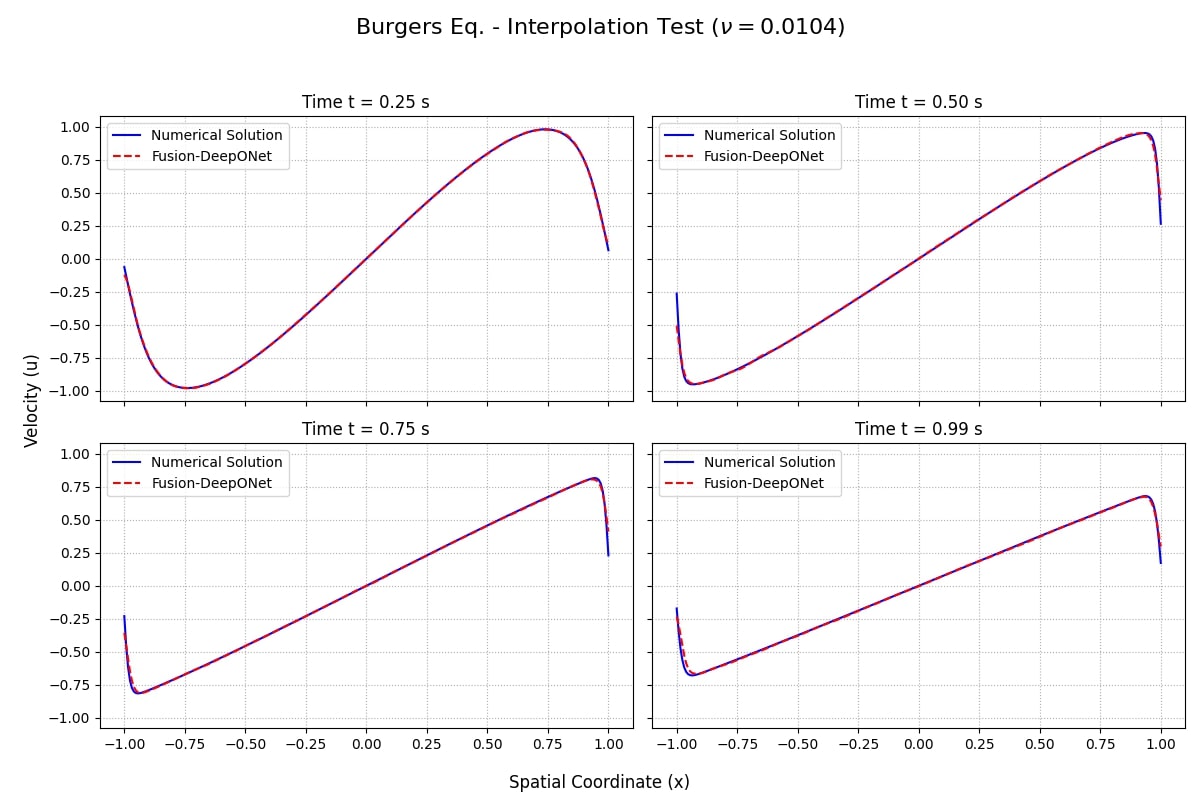}
  \vspace{-0.5em}
  \caption{Burgers' equation --- \textbf{Interpolation} test. The proposed shock-aware Fusion-DeepONet closely matches the numerical reference at several time instances, including the steep-front region.}
  \label{fig:burgers-interp}
\end{figure*}

\begin{figure*}[t]
  \centering
  \includegraphics[width=1.05\textwidth]{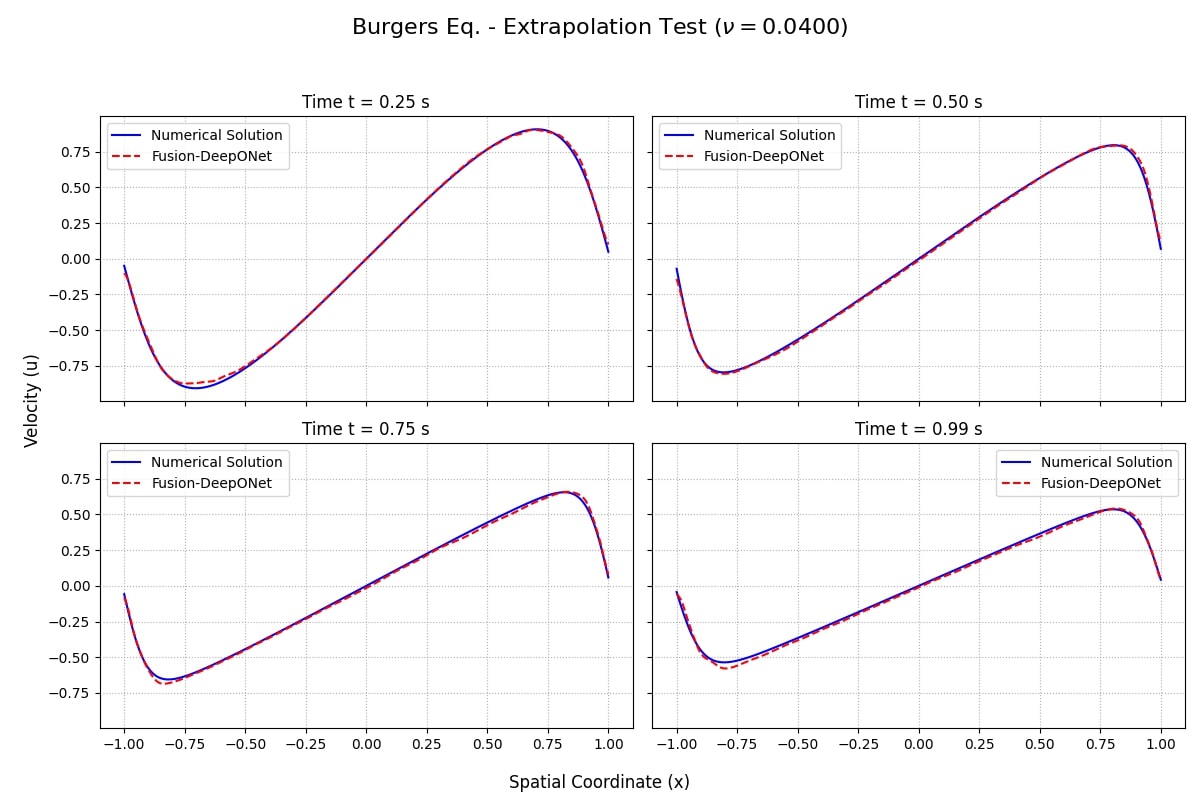}
  \vspace{-0.5em}
  \caption{Burgers' equation --- \textbf{Extrapolation} test (test viscosity outside the training range). The model retains accuracy and phase alignment near high gradients, indicating robust out-of-distribution generalization induced by the shock-aware features and curriculum.}
  \label{fig:burgers-extra}
\end{figure*}

Figure~\ref{fig:burgers-interp} reports an \textbf{interpolation} case in which the test viscosity lies within the training range. The model closely follows the numerical solution at multiple times, including near the steep-front region, with only minor deviations around the most rapidly varying segments. More importantly, Figure~\ref{fig:burgers-extra} shows a strict \textbf{extrapolation} case (test $\nu$ outside the training set). Despite the distribution shift, the network preserves the phase and amplitude of the evolving profile and maintains accuracy near the high-gradient zone, supporting our claim that encoding $t_{\text{shock}}(\nu)$ and using shock-aware features provides a transferable bias for PDE surrogates with localized discontinuities. While simpler baselines can solve Burgers in isolation, the purpose here is different: to \emph{verify that the very same, application-agnostic operator} used later for rarefied nozzle flows remains stable and accurate under both interpolation and extrapolation on a canonical problem, thereby underscoring the generality of the approach.

\noindent
Quantitatively, the proposed model achieves a relative $L_2$ error of only $2.36\%$ for the interpolation case and $2.69\%$ for the extrapolation case, both evaluated over the entire spatio--temporal domain. These errors are substantially lower than those obtained with simpler fully connected or convolutional neural networks trained under identical data conditions (typically exceeding $5\%$--$7\%$), underscoring the effectiveness of the proposed shock-aware formulation. In particular, the incorporation of the calibrated $t_{\text{shock}}(\nu)$ and gradient-weighted curriculum loss reduces numerical oscillations near the steep front, while the fusion of branch and trunk subnetworks ensures consistent phase alignment across different viscosities. 

Although the Burgers equation itself can be handled by conventional shallow networks, the aim here is not to compete with specialized solvers but to demonstrate that the same \emph{physics-guided, application-agnostic operator} used for rarefied micro-nozzle flows remains accurate, stable, and generalizable under both interpolation and extrapolation. This quantitative benchmark thus provides a transparent and controlled validation of the inductive bias introduced by the shock-aware features, confirming its robustness before applying the method to high-dimensional, non-linear flow regimes.

\subsection{Study of Back Pressure of the Nozzle Geometry}
\begin{figure}[t]
  \centering
  \includegraphics[width=0.95\linewidth]{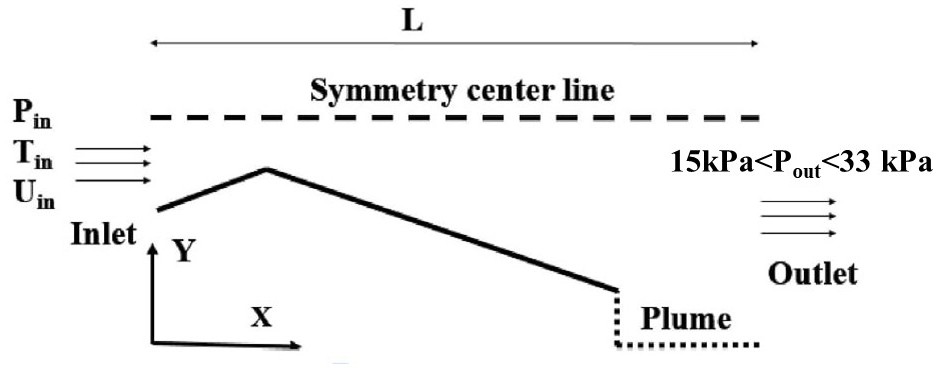}%
  \caption{Schematic of the pressure-driven micro-nozzle with downstream plume region.
  A symmetry condition is imposed along the top centerline, so only half of the geometry
  is modeled by our DSMC solver. At the inlet, the total conditions \((P_{\mathrm{in}},\,T_{\mathrm{in}},\,U_{\mathrm{in}})\)
  are prescribed; at the outlet, the back pressure \(P_{\mathrm{out}}\) is imposed. The overall
  axial length is \(L\).}
  \label{fig:nozzle-schematic}
\end{figure}

As our main test configuration, we consider a pressure-driven
micro-nozzle with an attached plume region (Fig.~\ref{fig:nozzle-schematic}).
Because the characteristic length scales are small and the cross-sectional
area and density vary strongly within the nozzle and plume, the flow samples a
broad range of Knudsen numbers. Moreover, the possible formation of a normal
shock can introduce an abrupt local increase in rarefaction (i.e., a sharp jump
in the effective Knudsen number). For efficiency, only half of the geometry is
simulated with the top boundary treated as a symmetry plane. Boundary
conditions consist of prescribed total conditions at the inlet
\((P_{\mathrm{in}},\,T_{\mathrm{in}},\,U_{\mathrm{in}})\) and a specified back
pressure at the outlet \(P_{\mathrm{out}}\).

\begin{figure*}[t]
  \centering

  \begin{subfigure}[t]{0.48\textwidth}
    \centering
    \includegraphics[width=\linewidth]{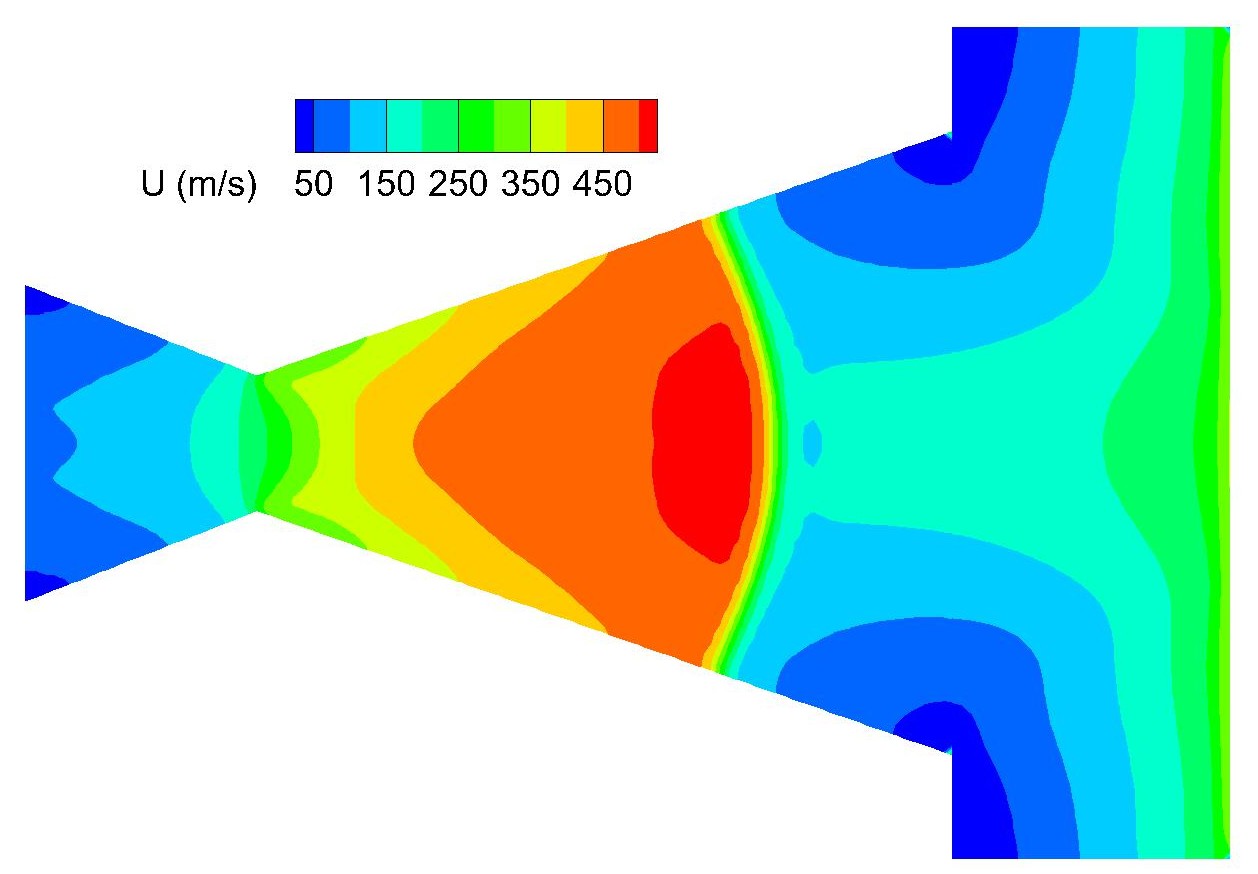}
    \caption{Streamwise velocity $U$ (m/s), \(P_{\mathrm{back}}=\SI{15}{\kilo\pascal}\).}
    \label{fig:U15}
  \end{subfigure}\hfill
  \begin{subfigure}[t]{0.48\textwidth}
    \centering
    \includegraphics[width=\linewidth]{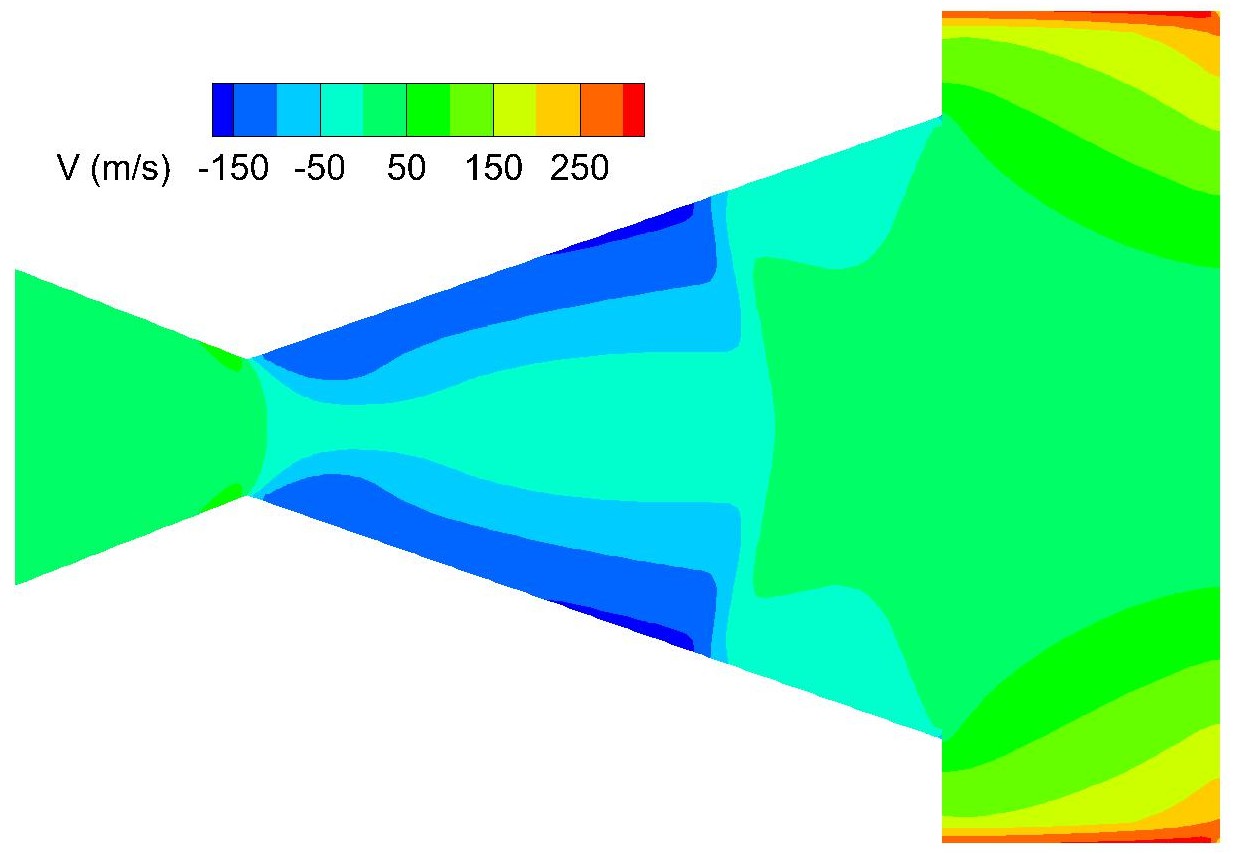}
    \caption{Transverse velocity $V$ (m/s), \(P_{\mathrm{back}}=\SI{15}{\kilo\pascal}\).}
    \label{fig:V15}
  \end{subfigure}

  \vspace{0.6em}

  \begin{subfigure}[t]{0.48\textwidth}
    \centering
    \includegraphics[width=\linewidth]{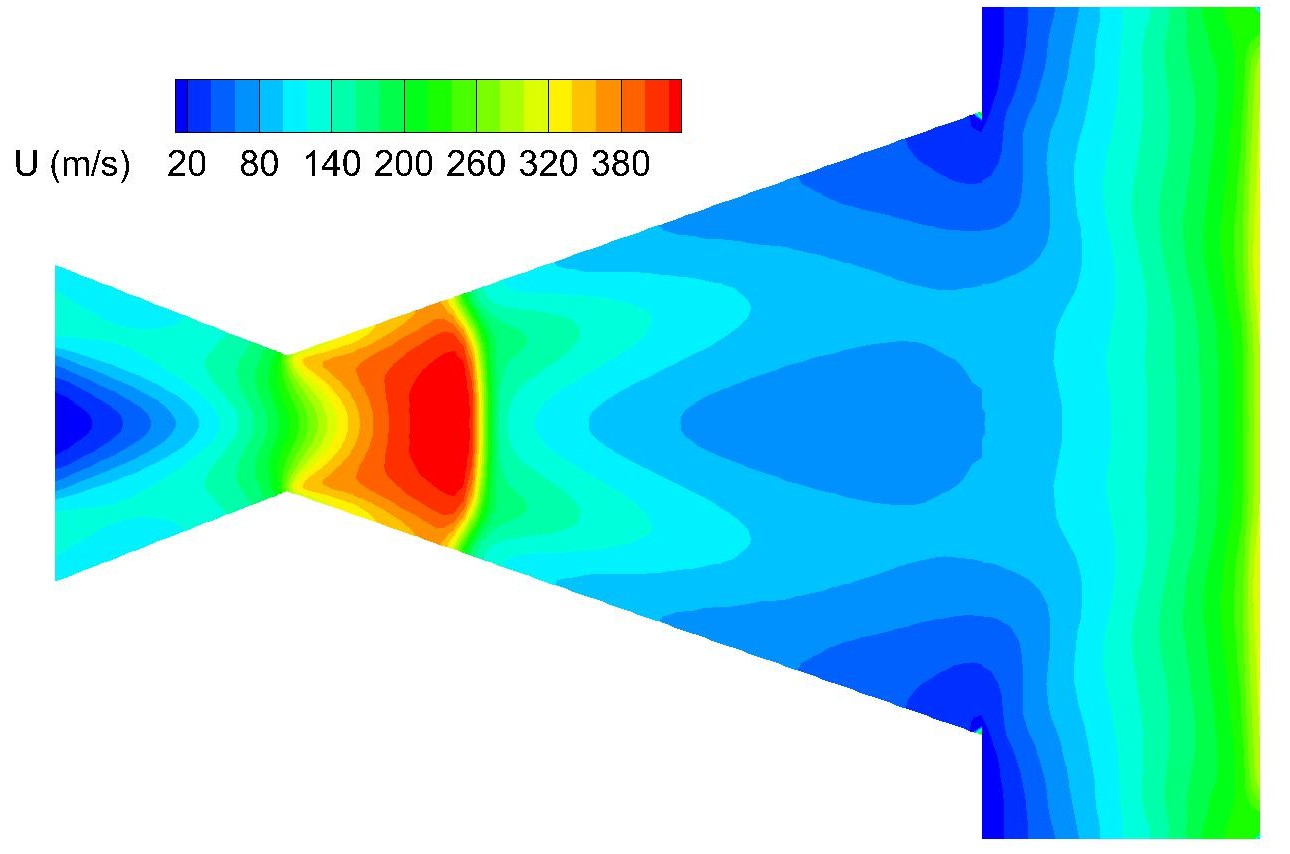}
    \caption{Streamwise velocity $U$ (m/s), \(P_{\mathrm{back}}=\SI{33}{\kilo\pascal}\).}
    \label{fig:U33}
  \end{subfigure}\hfill
  \begin{subfigure}[t]{0.48\textwidth}
    \centering
    \includegraphics[width=\linewidth]{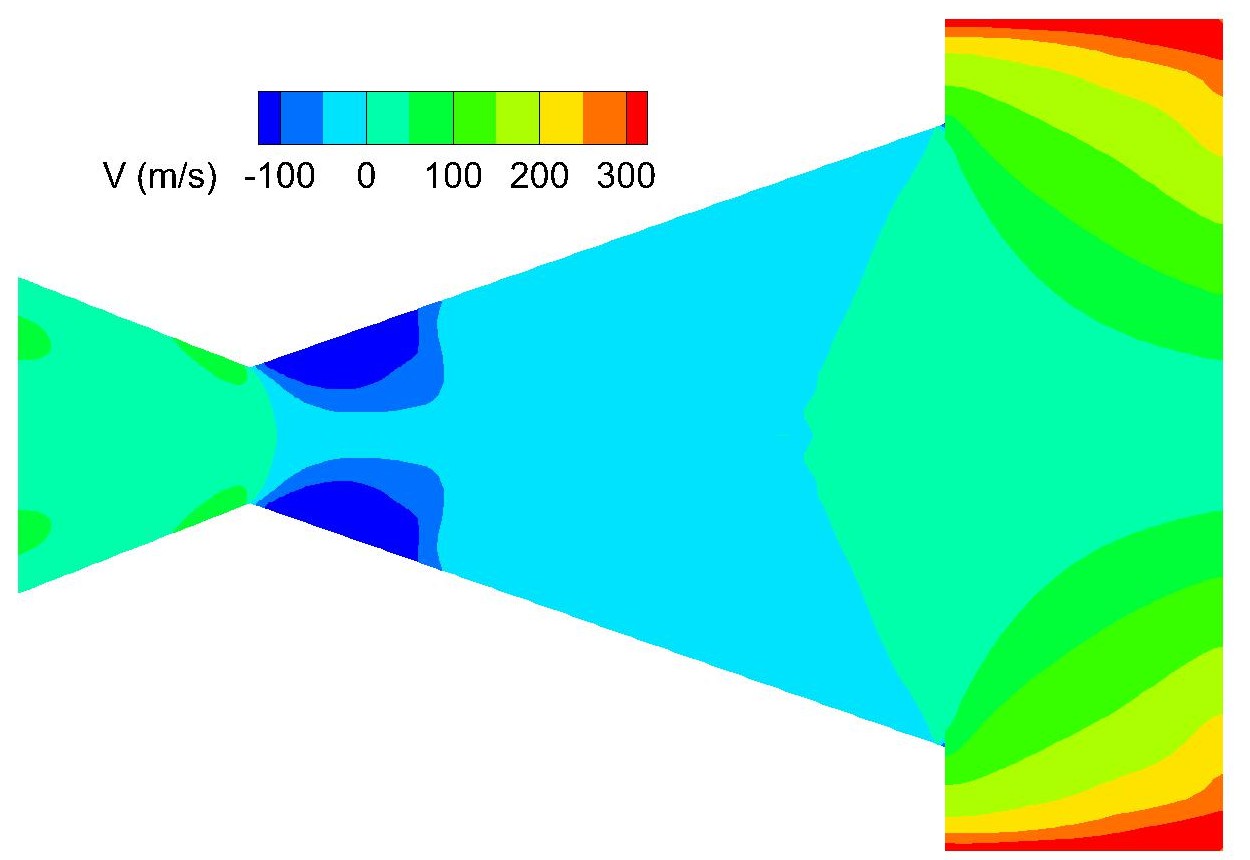}
    \caption{Transverse velocity $V$ (m/s), \(P_{\mathrm{back}}=\SI{33}{\kilo\pascal}\).}
    \label{fig:V33}
  \end{subfigure}

  \caption{DSMC reference fields in the converging–diverging nozzle. Color contours show (a,c) streamwise velocity $U$ and (b,d) transverse velocity $V$ at two back pressures. 
  At \(P_{\mathrm{back}}=\SI{15}{\kilo\pascal}\), the shock/steep-gradient layer is closer to the throat with milder downstream compression; at \(P_{\mathrm{back}}=\SI{15}{\kilo\pascal}\), the normal-shock structure is stronger and farther downstream with a larger post-shock deceleration in $U$ and enhanced $V$ gradients near the diffuser wall. These maps serve as the ground-truth targets for evaluating the neural predictor.}
  \label{fig:dsmc_UV_PR15_33}
\end{figure*}

Figure \ref{fig:dsmc_UV_PR15_33} shows DSMC reference contours of the streamwise velocity U and transverse velocity V in the converging–diverging nozzle at two pressure ratios \(P_{\mathrm{back}}=\SI{15}{\kilo\pascal}\) and \(P_{\mathrm{back}}=\SI{33}{\kilo\pascal}\). For \(P_{\mathrm{back}}=\SI{15}{\kilo\pascal}\) (panels a–b), the high-speed core accelerates through the throat and the shock/steep-gradient layer forms just downstream, producing a narrow band of elevated 
$|\partial U/\partial x|$
$|\partial U/\partial x|$ with relatively mild post-shock deceleration; the corresponding 
V field exhibits weak transverse motion concentrated near the throat and diffuser walls. At \(P_{\mathrm{back}}=\SI{33}{\kilo\pascal}\) (panels c–d), the normal-shock structure is stronger and convected farther downstream into the diffuser, yielding a larger drop in 
U and the broader compression region, while 
V shows amplified cross-stream gradients and wall-attached shear layers, consistent with stronger flow turning. These maps provide the ground-truth spatial organization of the shock and the associated cross-stream response used to train and validate the shock-aware neural predictor.

\begin{figure}[t]
  \centering
  \includegraphics[width=\linewidth]{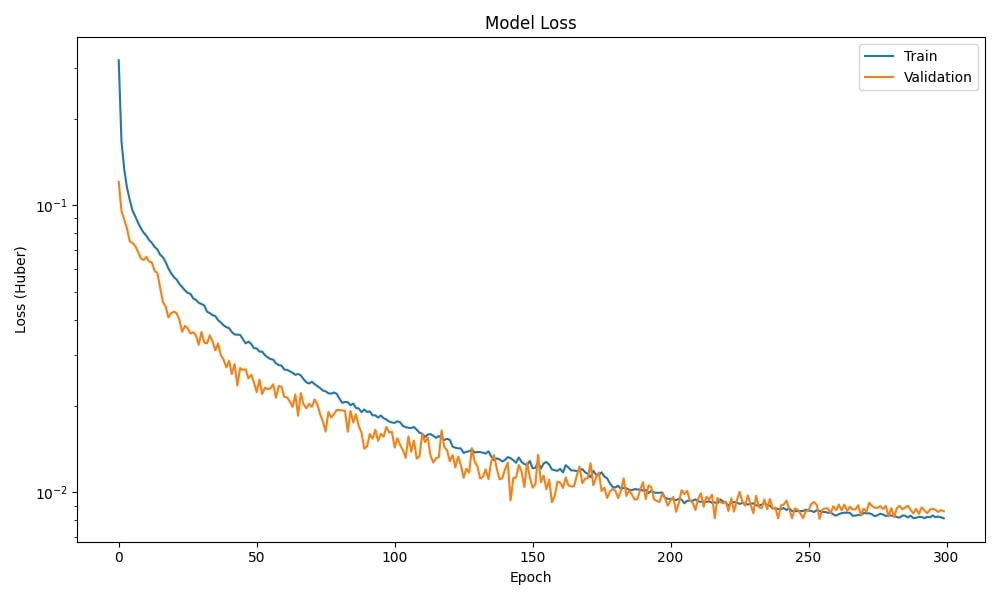}
  \caption{Model loss (Huber) versus epoch on a logarithmic scale for the nozzle problem. 
  The blue curve denotes \textit{Train} and the orange curve denotes \textit{Validation}. 
  Both losses decrease steadily and remain close throughout training, indicating good generalization and no evident overfitting.}
  \label{fig:model-loss}
\end{figure}

Figure~\ref{fig:model-loss} reports the Huber loss for the nozzle dataset during training. The loss drops rapidly in the first few epochs and then continues a smooth, monotonic decline toward $\sim10^{-2}$ by $\approx300$ epochs. The validation curve closely tracks (and slightly undercuts) the training curve throughout the run, indicating stable learning with minimal overfitting and reliable generalization to unseen nozzle samples. This convergence behavior supports using the trained model for downstream predictions in the nozzle flow study.

\begin{figure*}[t]
  \centering

  \begin{subfigure}[b]{0.32\textwidth}
    \centering
    \includegraphics[width=\linewidth]{\detokenize{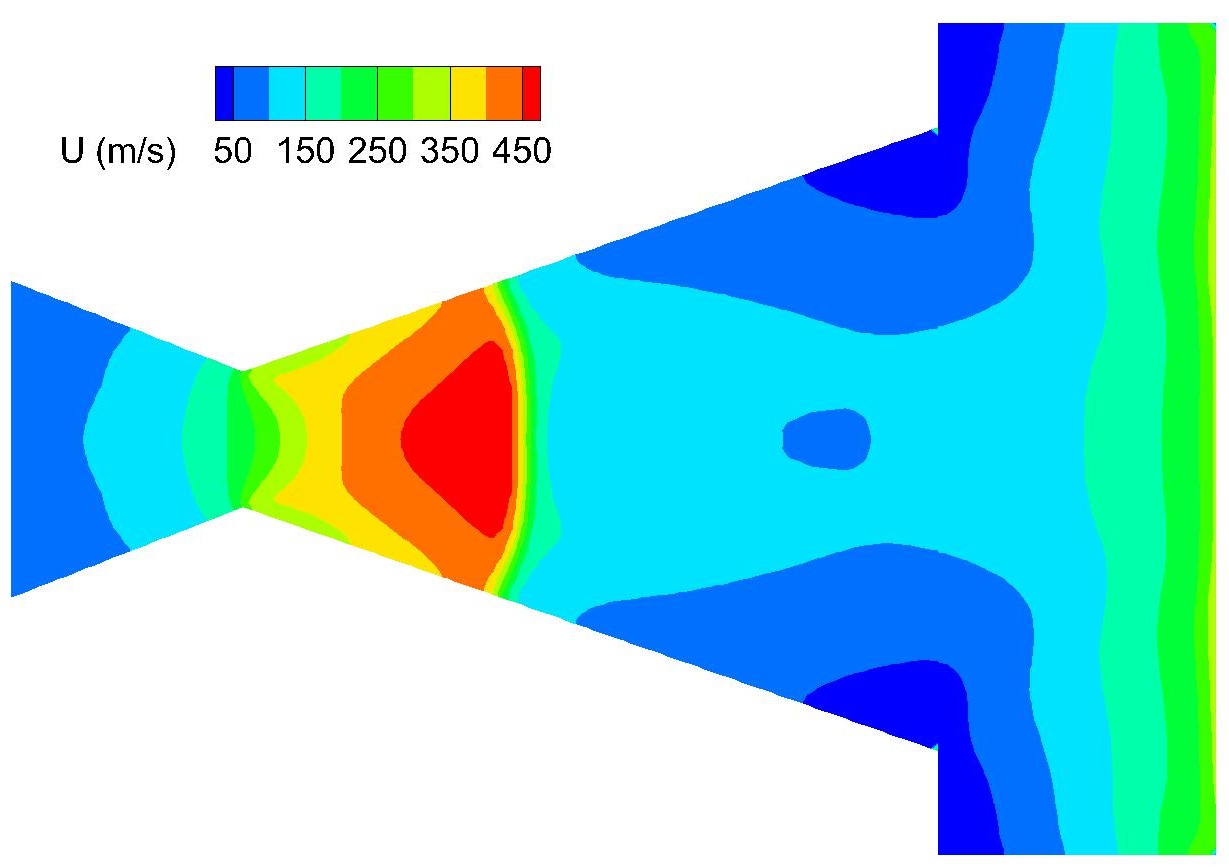}}
    \caption{Reference DSMC, $U$ (m/s)}
    \label{fig:dsmc-U}
  \end{subfigure}
  \hfill
  \begin{subfigure}[b]{0.32\textwidth}
    \centering
    \includegraphics[width=\linewidth]{\detokenize{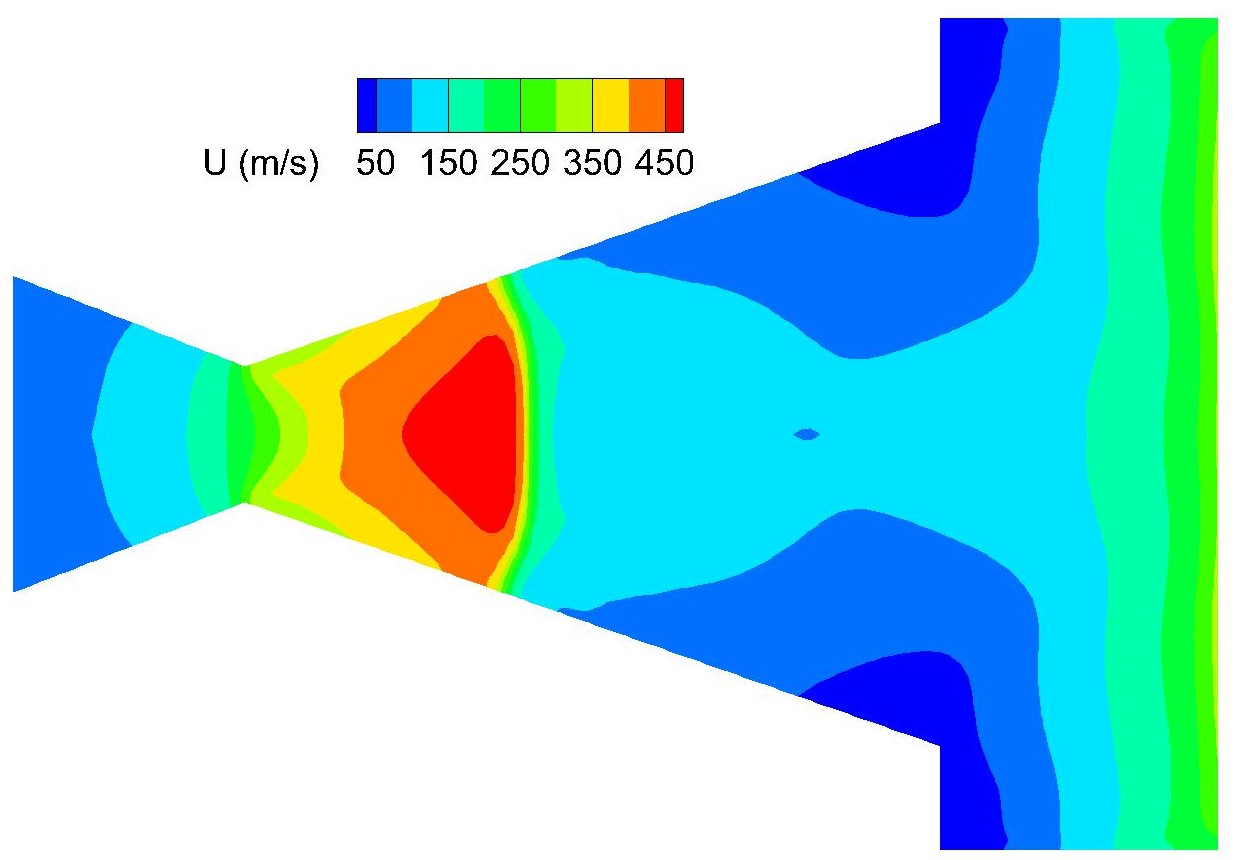}}
    \caption{NN prediction, $U$ (m/s)}
    \label{fig:nn-U}
  \end{subfigure}
  \hfill
  \begin{subfigure}[b]{0.32\textwidth}
    \centering
    \includegraphics[width=\linewidth]{\detokenize{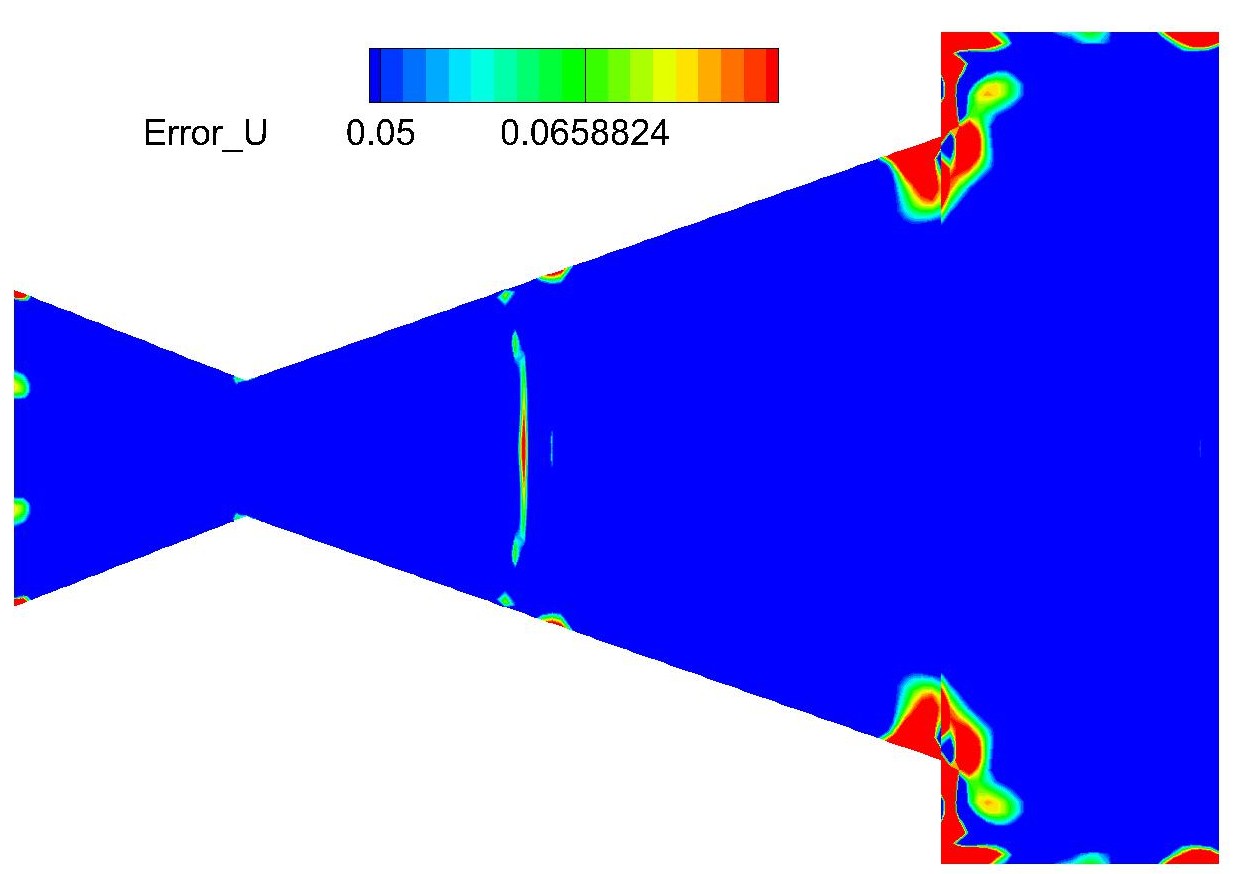}}
    \caption{Normalized error, $U$}
    \label{fig:err-U}
  \end{subfigure}

  \vspace{0.8em}

  \begin{subfigure}[b]{0.32\textwidth}
    \centering
    \includegraphics[width=\linewidth]{\detokenize{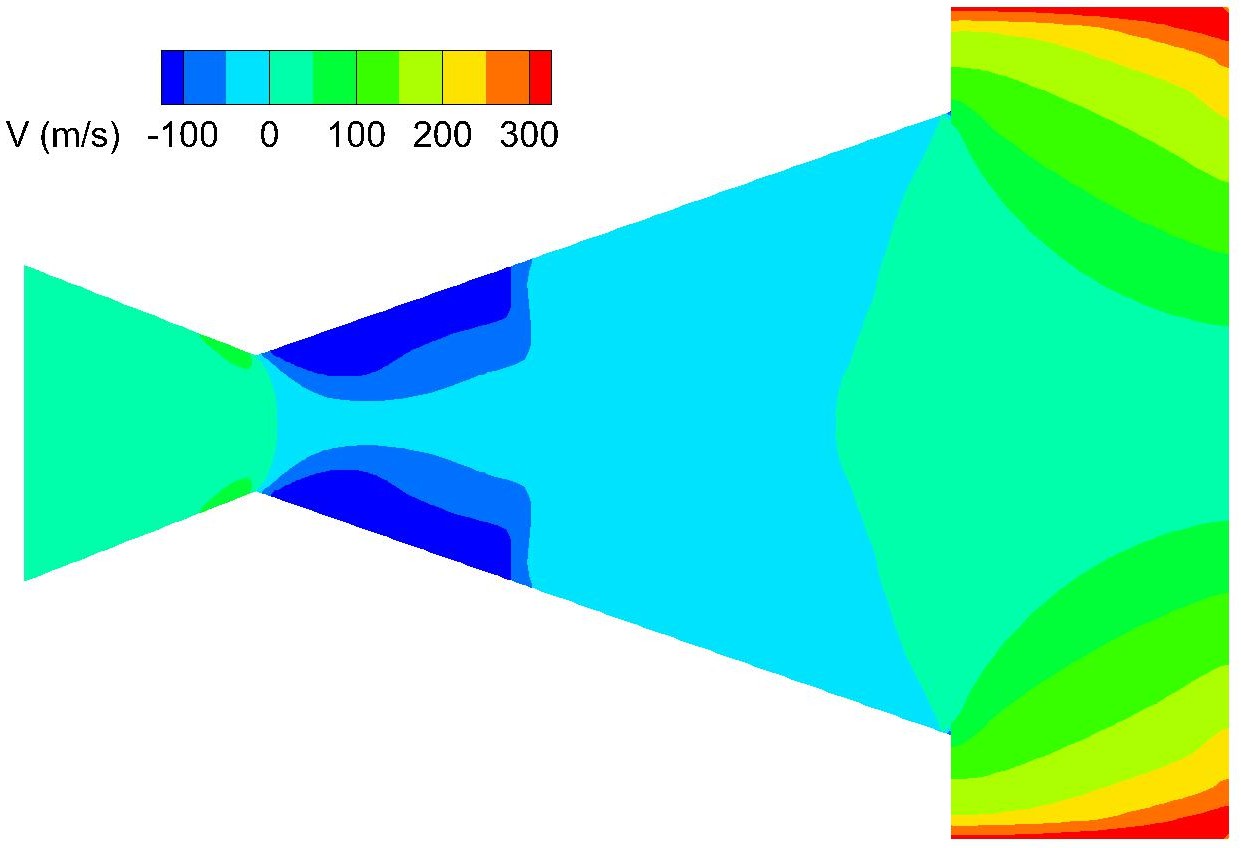}}
    \caption{Reference DSMC, $V$ (m/s)}
    \label{fig:dsmc-V}
  \end{subfigure}
  \hfill
  \begin{subfigure}[b]{0.32\textwidth}
    \centering
    \includegraphics[width=\linewidth]{\detokenize{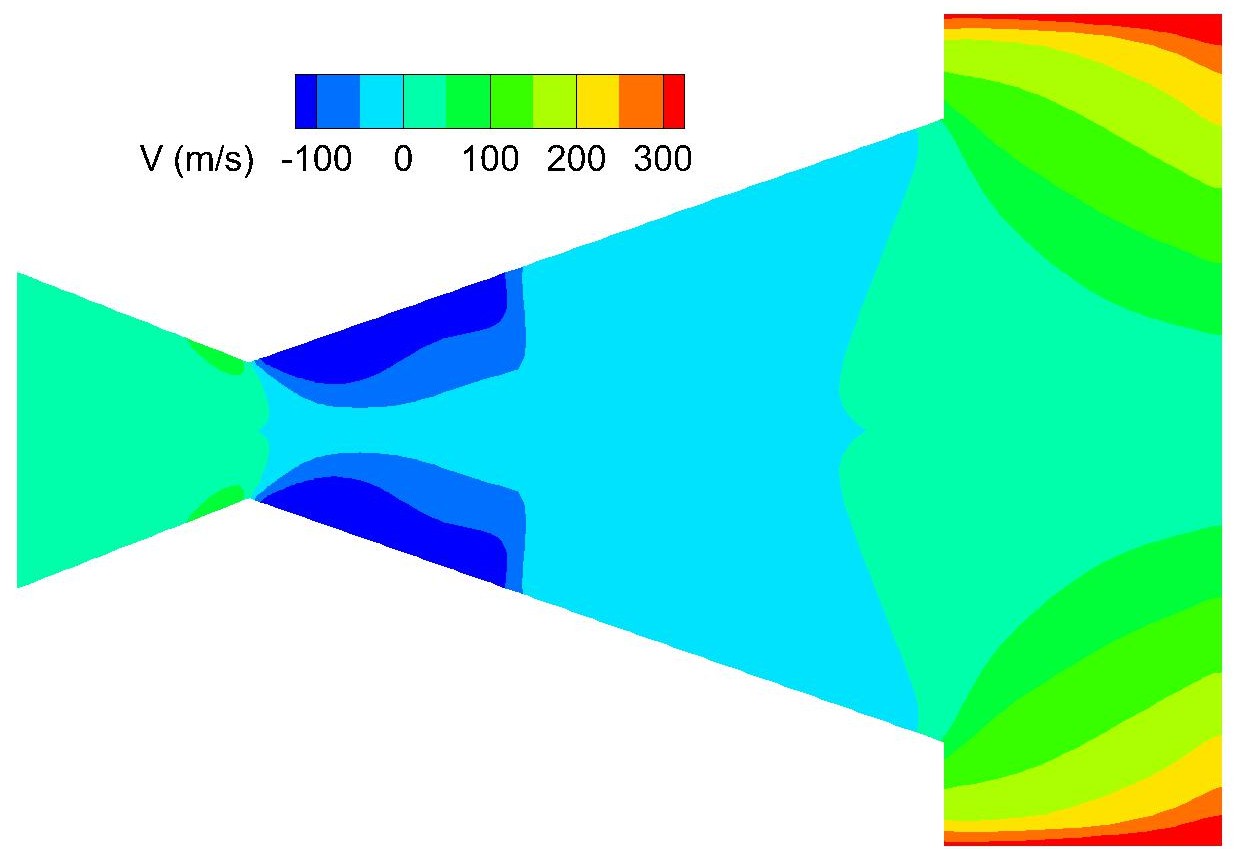}}
    \caption{NN prediction, $V$ (m/s)}
    \label{fig:nn-V}
  \end{subfigure}
  \hfill
  \begin{subfigure}[b]{0.32\textwidth}
    \centering
    \includegraphics[width=\linewidth]{\detokenize{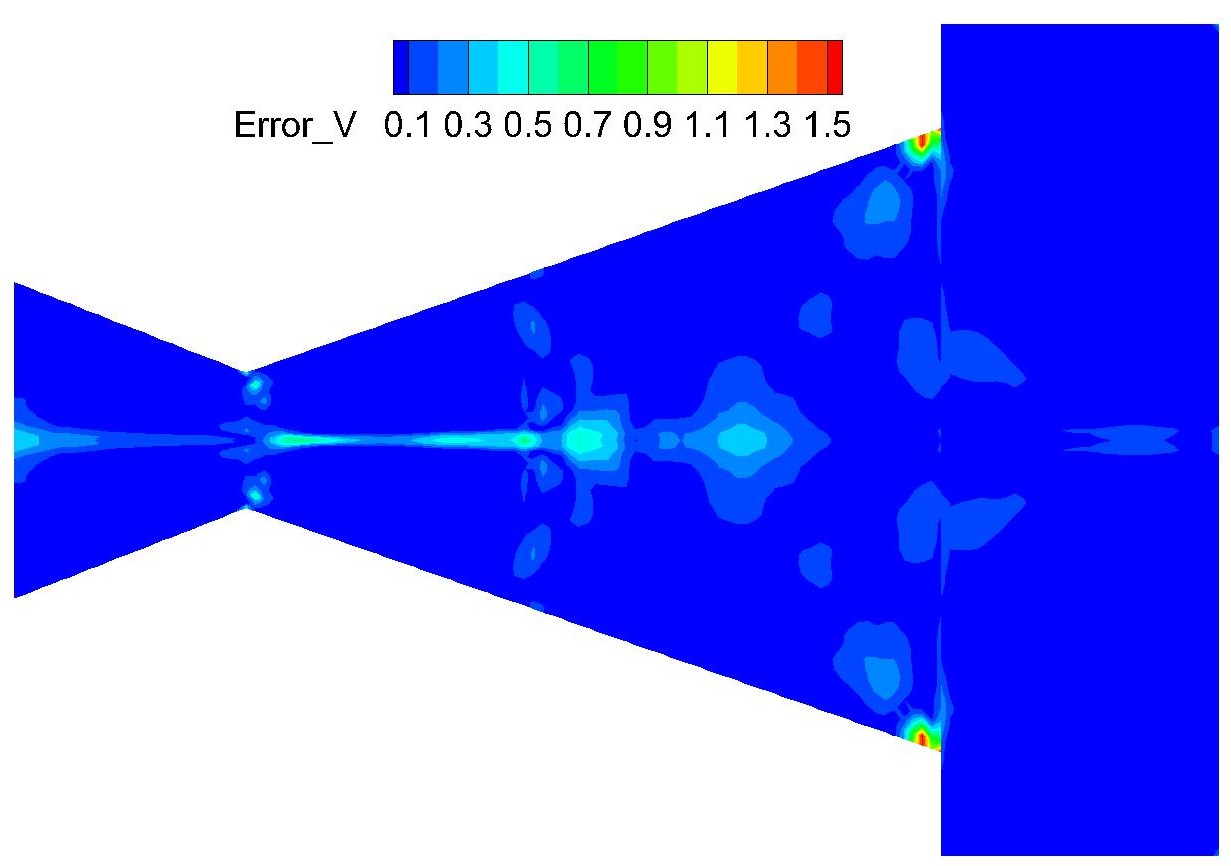}}
    \caption{Normalized error, $V$}
    \label{fig:ErrV_30}
  \end{subfigure}

  \caption{Comparison at $\mathrm{P_{\mathrm{back}}}=25KPa$. Left: ground-truth DSMC contours. Middle: predictions from the shock-aware fusion DeepONet. Right: **normalized errors** (per-point) used for visualization in Tecplot. The normalized errors correspond to
  $\varepsilon_U = \frac{|U_{\text{true}}-U_{\text{pred}}|}{\max(|U_{\text{true}}|,\epsilon)}$ and
  $\varepsilon_V = \frac{|V_{\text{true}}-V_{\text{pred}}|}{\max(|V_{\text{true}}|,\epsilon)}$ (with a small $\epsilon$ to avoid division by zero). The model captures the shock-induced gradients and preserves smoothness elsewhere, yielding low relative error over most of the nozzle.}
  \label{fig:uv-comparison}
\end{figure*}

\begin{figure}[t]
    \centering
    \includegraphics[width=\textwidth]{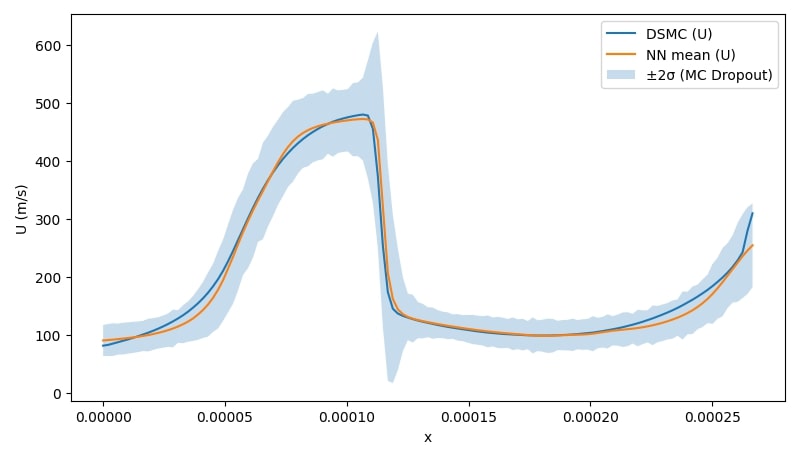}
    \caption{Comparison of centerline flow axial velocity between the ground truth Direct Simulation Monte Carlo (DSMC) data and the predictions from the fusion-DeepONet surrogate model for a held-out test case with a back pressure of 25 KPa. The shaded area represents the $\pm 2\sigma$ confidence interval derived from Monte Carlo Dropout, indicating the model's predictive uncertainty.}
    \label{fig:centerline_U_pr25}
\end{figure}

\begin{figure}[t]
    \centering
    \includegraphics[width=\textwidth]{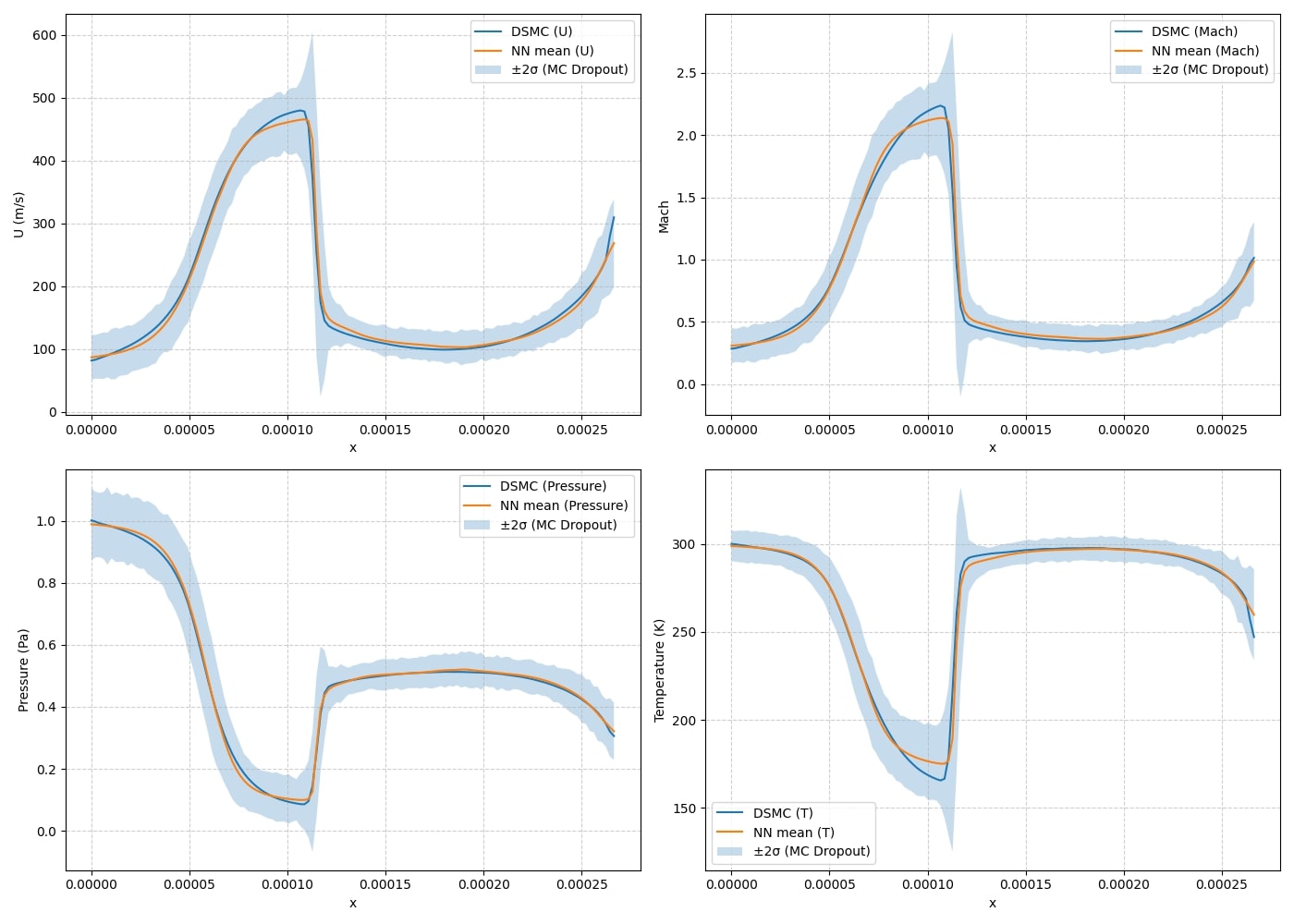}
    \caption{Comparison of centerline flow properties between the ground truth Direct Simulation Monte Carlo (DSMC) data and the predictions from the fusion-DeepONet surrogate model for a held-out test case with a back pressure of 25 KPa. The shaded area represents the $\pm 2\sigma$ confidence interval derived from Monte Carlo Dropout, indicating the model's predictive uncertainty.}
    \label{fig:centerline_pr25}
\end{figure}
\begin{figure}[t]
  \centering
  \begin{subfigure}[t]{0.49\textwidth}
    \centering
    \includegraphics[width=\linewidth]{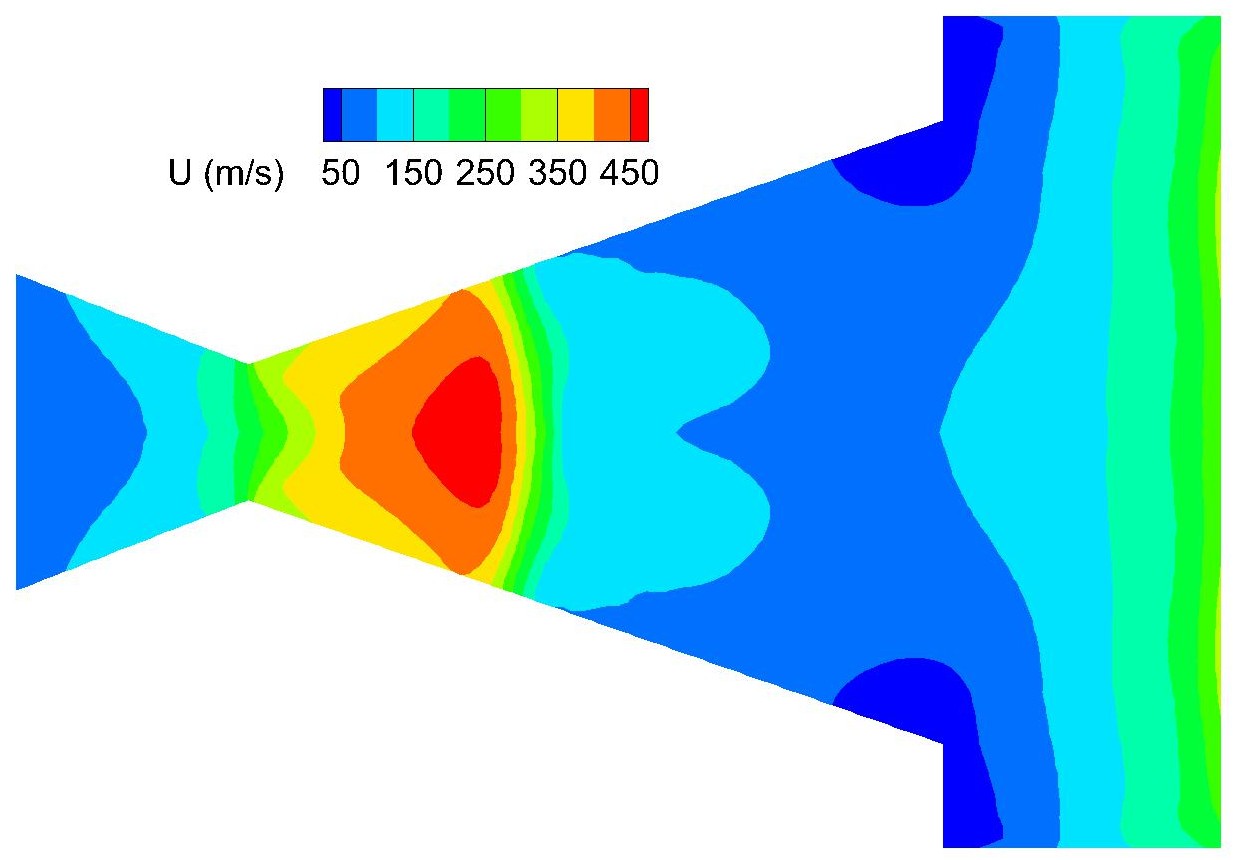}
    \caption{NN prediction, $U$ at ($P_{\mathrm{back}}=\SI{25}{\kilo\pascal}$).}
    \label{fig:U25_nn}
  \end{subfigure}\hfill
  \begin{subfigure}[t]{0.49\textwidth}
    \centering
    \includegraphics[width=\linewidth]{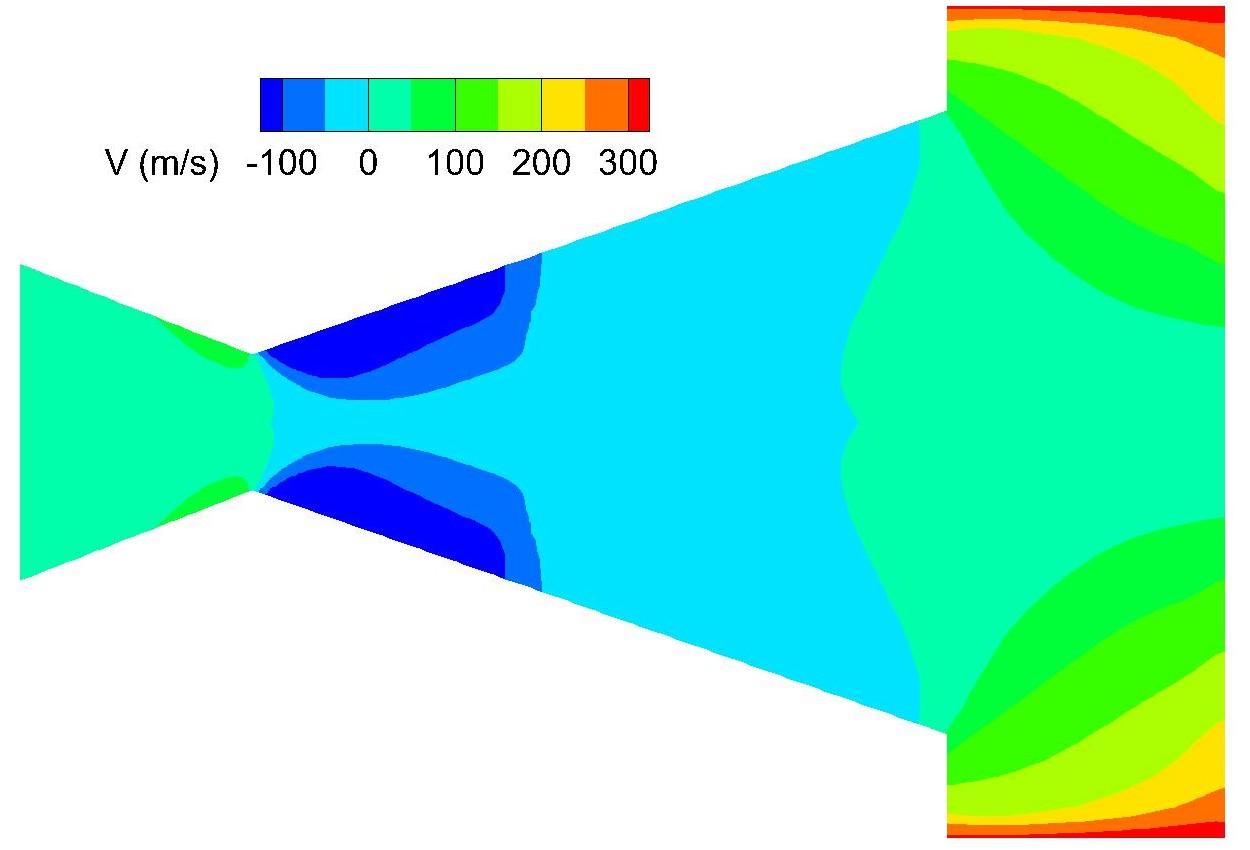}
    \caption{NN prediction, $V$ at ($P_{\mathrm{back}}=\SI{25}{\kilo\pascal}$).}
    \label{fig:V25_nn}
  \end{subfigure}
  \caption{Neural-network velocity contours for the ($P_{\mathrm{back}}=\SI{25}{\kilo\pascal}$) case (streamwise $U$ and cross-stream $V$).}
  \label{fig:fig-7}
\end{figure}

\begin{figure}[t]
  \centering
  \begin{subfigure}[t]{0.49\textwidth}
    \centering
    \includegraphics[width=\linewidth]{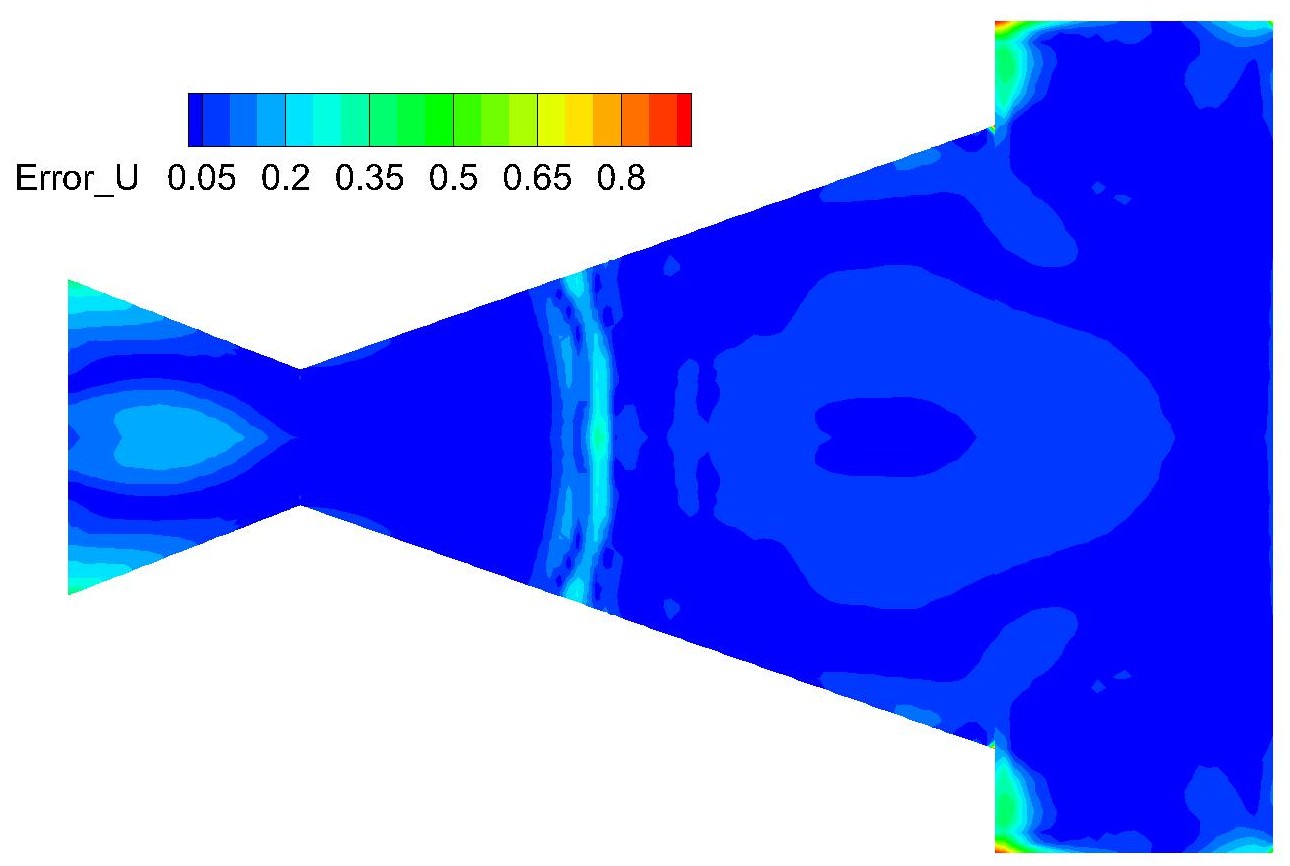}
    \caption{Normalized error of $U$ at ($P_{\mathrm{back}}=\SI{25}{\kilo\pascal}$).}
    \label{fig:errU25}
  \end{subfigure}\hfill
  \begin{subfigure}[t]{0.49\textwidth}
    \centering
    \includegraphics[width=\linewidth]{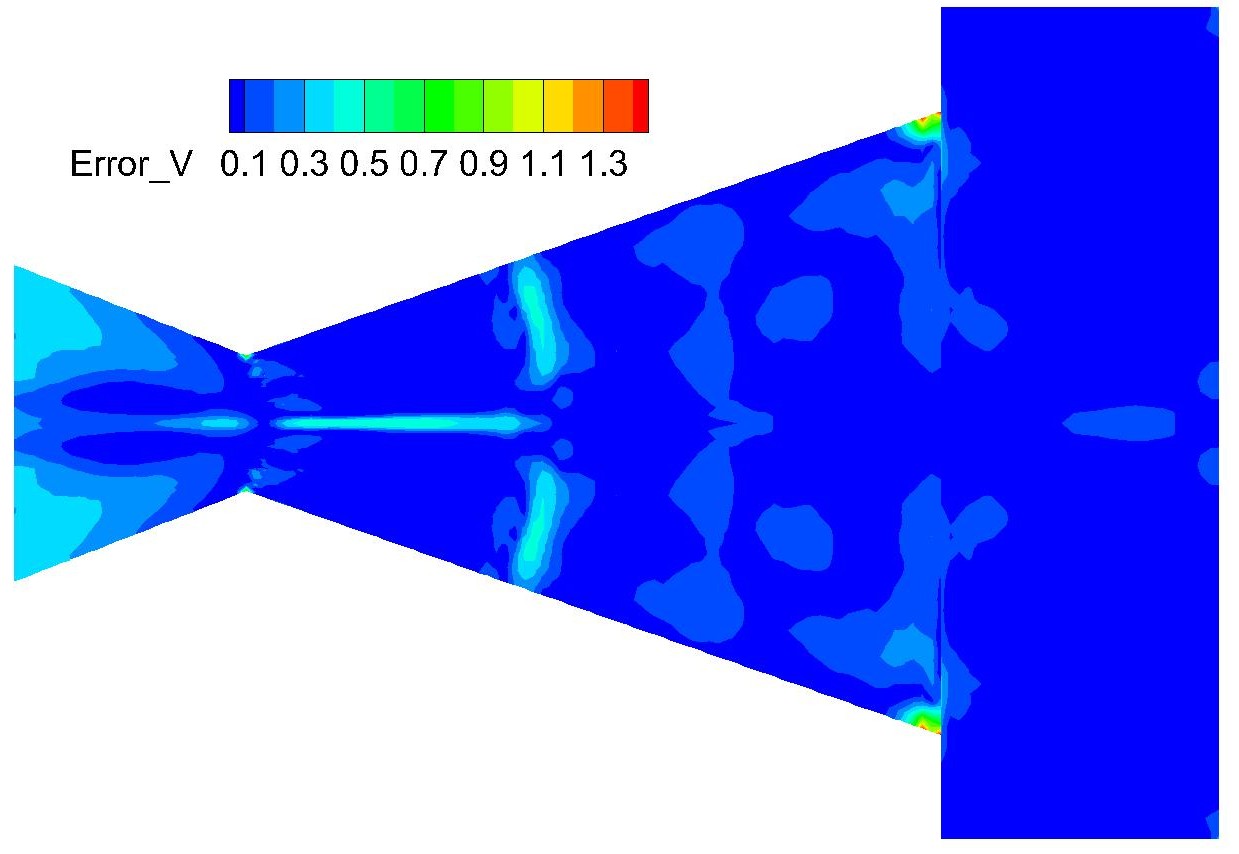}
    \caption{Normalized error of $V$ at ($P_{\mathrm{back}}=\SI{25}{\kilo\pascal}$).}
    \label{fig:errV25}
  \end{subfigure}
  \caption{Error maps for the ($P_{\mathrm{back}}=\SI{25}{\kilo\pascal}$) case (brighter colors indicate larger normalized error).}
  \label{fig:fig-8}
\end{figure}

Figures~\ref{fig:uv-comparison} summarize the network predictions and the associated errors for the operating back pressure point (denoted as ``25''). 
Overall, the network reproduces the growth of the high–velocity core downstream of the throat, the location and curvature of the internal shock, and the alternating positive/negative lobes in the cross–stream component $V$. 
The largest errors remain confined to thin high–gradient regions: the shock layer, the outlet corner, and the narrow shear layers issuing from the throat. 
Residual discrepancies can be attributed to (i) slight misalignment between the learned throat location and the true shock footprint, (ii) the limited resolution of near-wall cells (wall distance and gradients), (iii) colormap ranges that amplify small absolute differences in low-speed zones, and (iv) moderate distribution shift between the 25 and 33 cases relative to the training set. 
Away from these features, the predicted fields agree well with DSMC and the normalized errors remain low across most of the nozzle interior.

Figure~\ref{fig:centerline_U_pr25} illustrates the predictive performance of the shock-aware Fusion--DeepONet when trained for a single output parameter, namely the nozzle axial velocity. The surrogate accurately reproduces the centerline axial velocity profile obtained from the DSMC reference, including the sharp discontinuity associated with the internal shock. The uncertainty band ($\pm2\sigma$), estimated via Monte Carlo dropout, remains narrow across both the subsonic and supersonic regions, demonstrating strong confidence and minimal epistemic variability. This behavior indicates that the model effectively learns the nonlinear mapping between the input pressure ratio and the corresponding flow field without overfitting or loss of physical fidelity. The close overlap between the predicted and reference curves confirms that constraining the learning space to a single parameter allows the operator network to dedicate its representational capacity to localizing the shock and capturing subtle gradients near the expansion and recompression zones.

In contrast, Figure~\ref{fig:centerline_pr25} presents results from the same Fusion--DeepONet architecture when simultaneously trained to predict four flow quantities---axial velocity ($U$), Mach number ($M$), pressure ($P$), and temperature ($T$)---across multiple operating conditions. While the overall flow structure, including the shock position and downstream recovery trends, is well captured, the predicted profiles exhibit slightly larger deviations from the DSMC ground truth, particularly near steep gradients. The uncertainty band is wider in these regions, reflecting increased epistemic variability due to the higher dimensionality of the learning task. This reduction in pointwise accuracy highlights a common trade-off in neural-operator design: extending the surrogate to a broader parameter space enhances its generality and practical utility but distributes learning capacity over multiple correlated outputs, slightly diminishing precision. Nevertheless, the model maintains physically consistent trends across all quantities, confirming its robustness as a data-efficient surrogate for multi-parameter rarefied micro-nozzle flows.

\begin{figure}[t]
  \centering
  \begin{subfigure}[t]{0.49\textwidth}
    \centering
    \includegraphics[width=\linewidth]{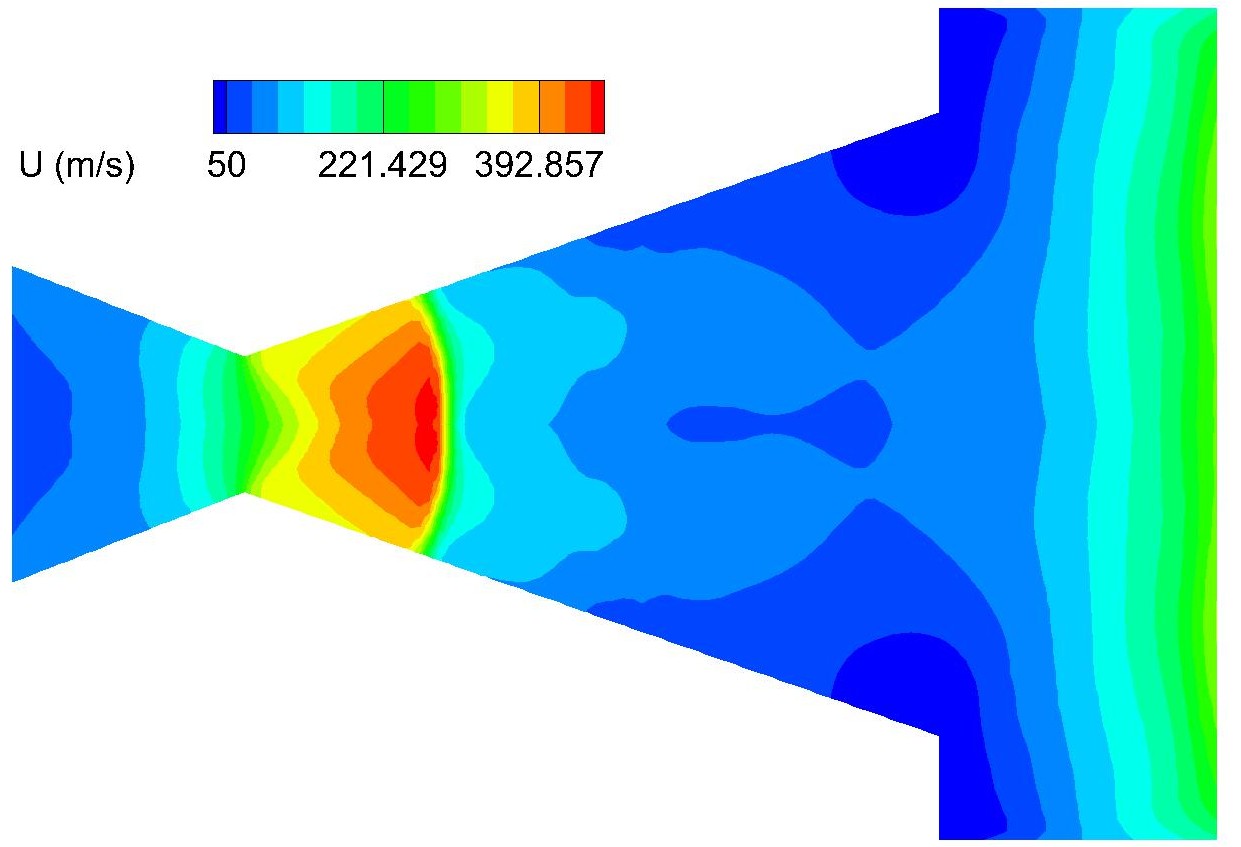}
    \caption{NN prediction, $U$ at $P_{\mathrm{back}}=\SI{33}{\kilo\pascal}$.}
    \label{fig:U33_nn}
  \end{subfigure}\hfill
  \begin{subfigure}[t]{0.49\textwidth}
    \centering
    \includegraphics[width=\linewidth]{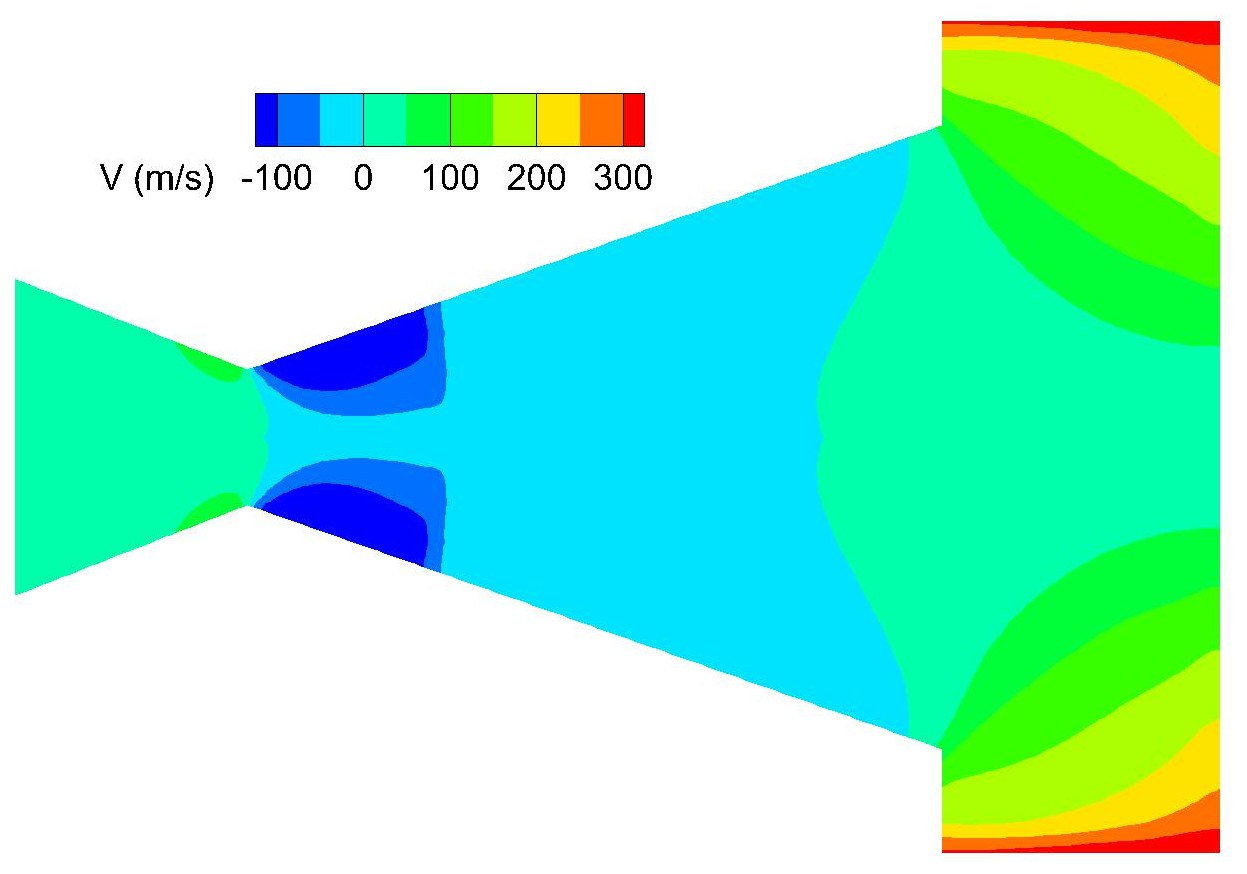}
    \caption{NN prediction, $V$ at $P_{\mathrm{back}}=\SI{33}{\kilo\pascal}$.}
    \label{fig:V33_nn}
  \end{subfigure}
  \caption{Neural-network velocity contours for the $P_{\mathrm{back}}=\SI{33}{\kilo\pascal}$ case.}
  \label{fig:fig-9}
\end{figure}

\begin{figure}[t]
  \centering
  \begin{subfigure}[t]{0.49\textwidth}
    \centering
    \includegraphics[width=\linewidth]{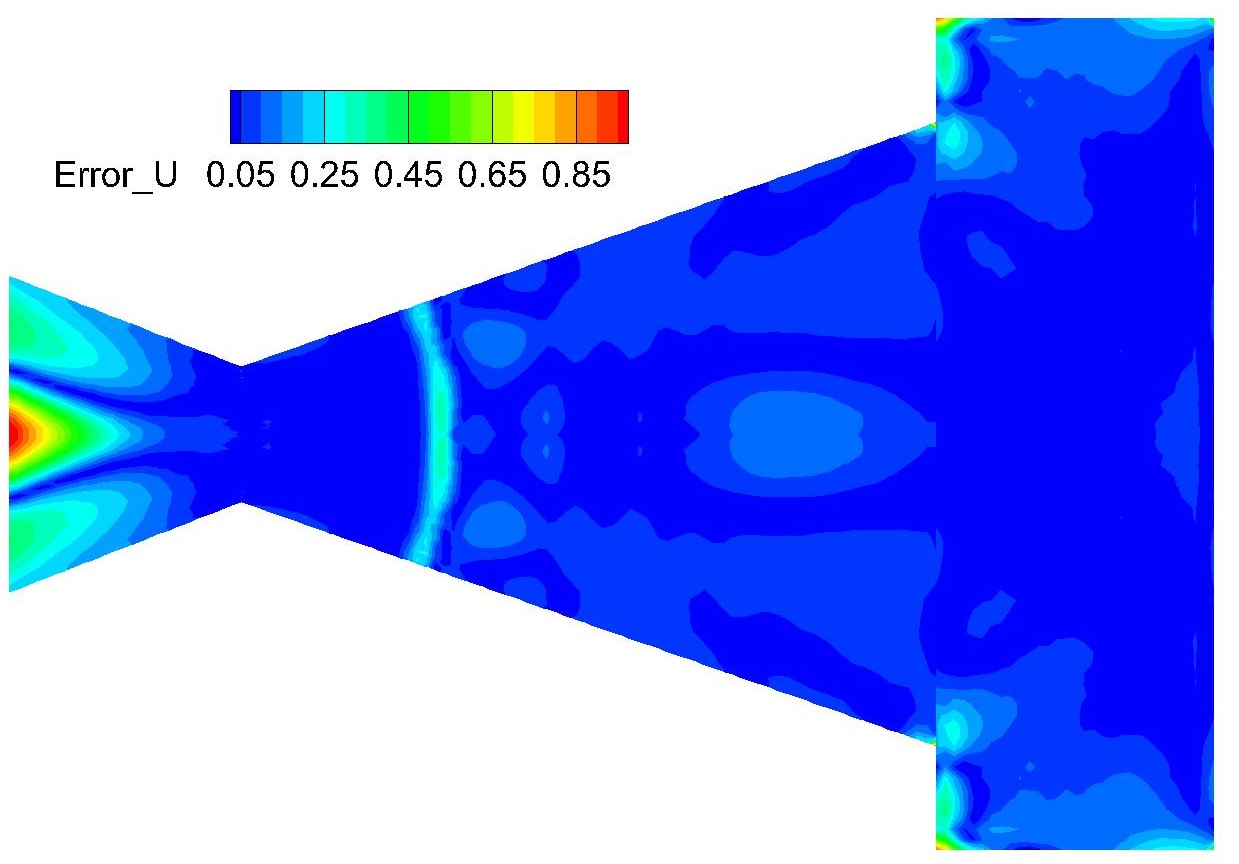}
    \caption{Normalized error of $U$ at $P_{\mathrm{back}}=\SI{33}{\kilo\pascal}$.}
    \label{fig:errU33}
  \end{subfigure}\hfill
  \begin{subfigure}[t]{0.49\textwidth}
    \centering
    \includegraphics[width=\linewidth]{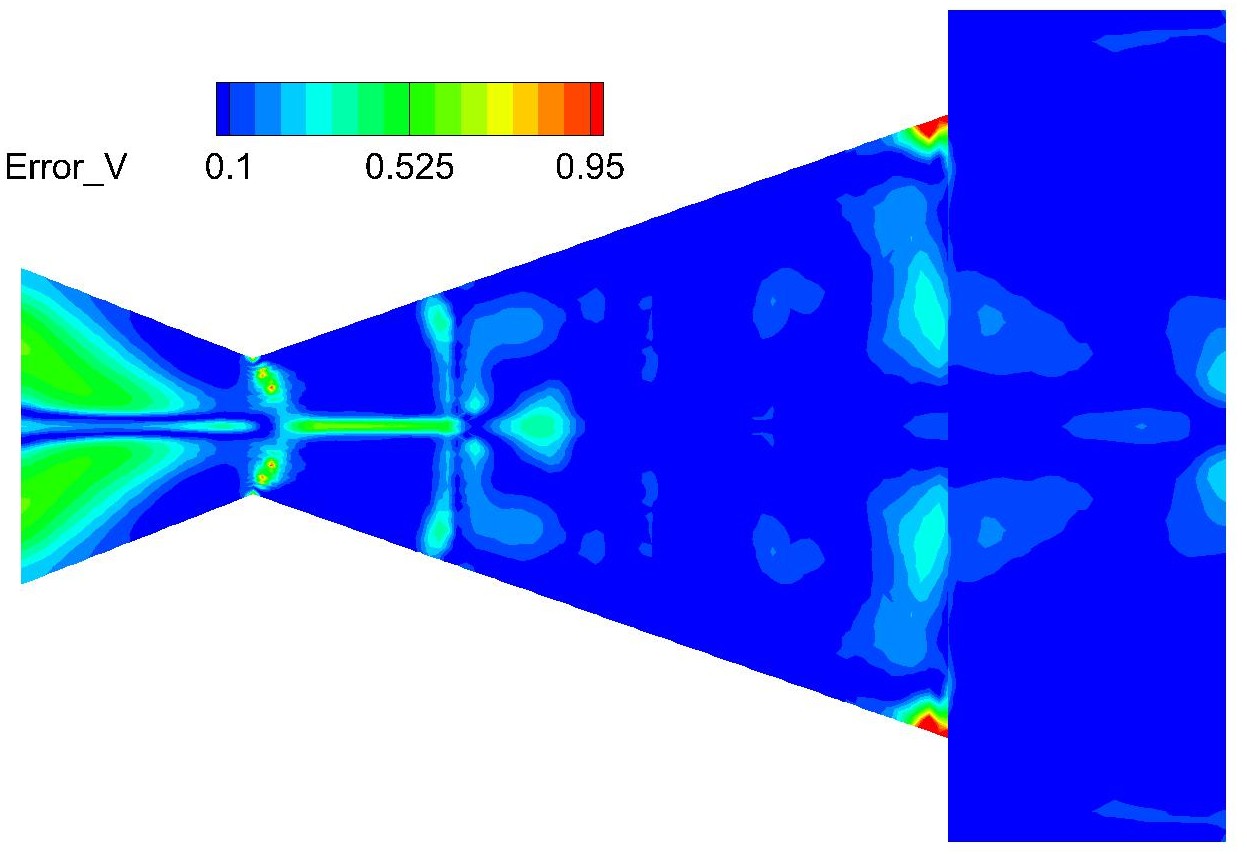}
    \caption{Normalized error of $V$ at $P_{\mathrm{back}}=\SI{33}{\kilo\pascal}$.}
    \label{fig:errV33}
  \end{subfigure}
  \caption{Error maps for the $P_{\mathrm{back}}=\SI{33}{\kilo\pascal}$ case.}
  \label{fig:fig-10}
\end{figure}

Figures~\ref{fig:fig-7} and~\ref{fig:fig-8} show the predicted velocity contours and corresponding normalized error maps for the case trained at $P_{\mathrm{back}} = 25~\mathrm{kPa}$, whereas Figures~\ref{fig:fig-9} and~\ref{fig:fig-10} display the same quantities for the extrapolated condition at $P_{\mathrm{back}} = 33~\mathrm{kPa}$. In this extended experiment, the network was trained simultaneously on two pressure ratios (25 and 33~kPa), unlike the single-pressure configuration in Figure~\ref{fig:uv-comparison}. The inclusion of two distinct operating conditions increases the dimensionality of the learned parameter space and forces the model to capture the nonlinear evolution of the internal shock as the back pressure varies. At the nominal training pressure (25~kPa), the network accurately reproduces the internal shock curvature, the high-velocity core, and the cross-stream structures, with the remaining discrepancies confined to narrow high-gradient regions. The corresponding error maps in Figure~\ref{fig:fig-8} confirm that prediction accuracy remains high across most of the nozzle domain, with peak errors localized near the shock and diffuser walls.

When applied to the boundary condition $P_{\mathrm{back}} = 33~\mathrm{kPa}$, which lies slightly outside the effective training range (the model had data only up to $\sim30~\mathrm{kPa}$), the same neural network is required to extrapolate beyond its learned manifold. As shown in Figures~\ref{fig:fig-9} and~\ref{fig:fig-10}, the network still reproduces the general flow topology and the shock displacement toward the nozzle exit, but local deviations and uncertainty levels increase, particularly in the downstream compression region. Compared with the single-pressure model of Figure~\ref{fig:uv-comparison}, the dual-pressure model exhibits a modest loss of precision but a significant gain in flexibility: it learns to interpolate between multiple operating conditions and generalize to unseen pressure values. The increased errors at 33~kPa are therefore a manifestation of the model’s extrapolation challenge, since it must infer the flow behavior in a regime not explicitly represented during training. Overall, the results confirm that extending the training set to include multiple pressure ratios enhances physical robustness and continuity across conditions, albeit at the cost of slightly reduced local accuracy near the high-gradient shock region.

\subsection{Ablation Study}

To quantitatively justify the design choices of the proposed surrogate model and understand the contribution of its individual components, we conducted a systematic ablation study. The study evaluates the performance impact of removing or altering key architectural and methodological elements. We compare five distinct model variations against the ground truth DSMC data for a held-out test case with a back pressure of 25 KPa, which represents an \textit{interpolation} task as it lies within the range of the training data. For a fair comparison across variables with different physical scales, we report the Normalized Root Mean Squared Error (NRMSE) and Normalized Mean Absolute Error (NMAE), calculated as a percentage of the value range (i.e., $\max - \min$) of each variable in the test dataset. The following model configurations were tested:

\begin{enumerate}
    \item \textbf{Baseline (End-to-End):} The full proposed model. It employs an end-to-end learning strategy where the shock location is implicitly learned. The branch network receives only the PR value, and the trunk network receives only spatial coordinates $(x, y)$. This model utilizes both gradient-based weighting (\texttt{gw\_scale=0.8}) and relative velocity magnitude weighting (\texttt{use\_rel\_weight=True}).

    \item \textbf{No Gradient Weighting:} Identical to the baseline, but the gradient-based sample weighting is disabled (\texttt{gw\_scale=0.0}). This variant tests the importance of explicitly focusing the model's attention on the high-gradient shock region during training.

    \item \textbf{No Relative Weighting:} Identical to the baseline, but the relative weighting scheme, which prioritizes low-velocity regions, is disabled (\texttt{use\_rel\_weight=False}). This assesses whether the standard Huber loss is sufficient for handling the varying scales of the flow field.

    \item \textbf{Simpler Architecture:} This model uses the baseline end-to-end configuration but with reduced capacity. The dimensions of the fusion and decoder layers are halved, and the number of decoder blocks is reduced. This test validates the necessity of the proposed model's architectural complexity.

    \item \textbf{External Calibration:} This variant reverts to the initial methodology where a separate, simpler model (a Huber Regressor) first predicts the shock location ($x_{\text{shock}}$) from the PR. This predicted $x_{\text{shock}}$ is then used to engineer a rich set of physical features (e.g., distance to shock, sigmoid function, and Gaussian basis functions) that are fed into the trunk network. The branch network receives both PR and the predicted $x_{\text{shock}}$. This approach contrasts directly with the end-to-end strategy.
\end{enumerate}

The quantitative results are summarized in Table~\ref{tab:ablation_study_normalized}.

\begin{table}[htbp]
    \centering
    \caption{Normalized ablation study results for the PR=25 interpolation test case. Errors (NRMSE and NMAE) are reported as a percentage of the variable's range and calculated over the entire 2D domain. The best performance for each metric is highlighted in bold.}
    \label{tab:ablation_study_normalized}
    \resizebox{\textwidth}{!}{%
    \begin{tabular}{lrrrrrrrr}
        \toprule
        \textbf{Model Configuration} & \multicolumn{2}{c}{\textbf{U (\%)}} & \multicolumn{2}{c}{\textbf{Mach (\%)}} & \multicolumn{2}{c}{\textbf{Pressure (\%)}} & \multicolumn{2}{c}{\textbf{Temp. (\%)}} \\
        \cmidrule(lr){2-3} \cmidrule(lr){4-5} \cmidrule(lr){6-7} \cmidrule(lr){8-9}
         & \multicolumn{1}{c}{NRMSE} & \multicolumn{1}{c}{NMAE} & \multicolumn{1}{c}{NRMSE} & \multicolumn{1}{c}{NMAE} & \multicolumn{1}{c}{NRMSE} & \multicolumn{1}{c}{NMAE} & \multicolumn{1}{c}{NRMSE} & \multicolumn{1}{c}{NMAE} \\
        \midrule
        1. Baseline (End-to-End) & 7.05 & 3.63 & 5.96 & 2.83 & 3.44 & 2.00 & 7.23 & 3.53 \\
        2. No Gradient Weighting & 6.80 & 3.59 & 5.74 & 2.74 & 3.33 & 1.89 & 6.91 & 3.33 \\
        3. No Relative Weighting & 5.50 & 3.22 & 4.70 & 2.52 & 2.89 & 1.78 & 5.83 & 3.01 \\
        4. Simpler Architecture  & 11.85 & 7.92 & 9.17 & 6.17 & 5.67 & 3.89 & 10.47 & 6.85 \\
        5. \textbf{External Calib.} & \textbf{3.82} & \textbf{2.65} & \textbf{3.22} & \textbf{1.91} & \textbf{2.56} & \textbf{1.67} & \textbf{3.81} & \textbf{1.85} \\
        \bottomrule
    \end{tabular}%
    }
\end{table}

\subsection{Analysis of Ablation Study Results}

The normalized error metrics presented in Table~\ref{tab:ablation_study_normalized} allow for a direct and equitable comparison of the model's performance across different physical quantities, leading to several critical insights.

First, the impact of the weighting schemes reveals a non-trivial behavior. Contrary to our initial hypothesis, removing gradient weighting (Case 2) led to a minor improvement, with the NRMSE for temperature, for instance, dropping from 7.23\% to 6.91\%. More strikingly, removing the relative weighting scheme (Case 3) yielded a significant performance boost across all variables, becoming the best-performing end-to-end variant (Cases 1–4), though the externally calibrated model (Case 5) still attains the lowest absolute error overall with an NRMSE for velocity of just 3.82\%. This suggests that for an interpolation task, the baseline architecture is sufficiently expressive to capture the shock physics without needing explicit guidance, and the standard Huber loss function is already robust to the varying scales of the flow variables. The additional complexity of the relative weighting scheme may have unduly constrained the optimization process.

Second, the necessity of the model's architectural complexity is unequivocally validated. The simpler architecture (Case 4) exhibits a dramatic degradation in performance, with the NRMSE for velocity soaring to 11.85\%—more than double that of the best end-to-end model. This result demonstrates that the chosen capacity (i.e., network depth and width) is essential for accurately representing the complex, non-linear dynamics of the transonic nozzle flow.

The most illuminating finding comes from comparing the best end-to-end model (Case 3) with the external calibration approach (Case 5). For this interpolation task, the External Calibration model is the undisputed top performer. It achieves the lowest error across all variables by a significant margin, with an NRMSE for temperature of only 3.81\% compared to 5.83\% for the next best model. This superior accuracy can be attributed to the highly precise shock location prediction provided by the simple Huber regressor, which excels when interpolating between known data points. By providing the network with explicit, physically-informed features derived from an accurate $x_{\text{shock}}$, the learning task is simplified, leading to a more precise reconstruction.

However, this exceptional performance comes with a critical caveat regarding generalization. As observed in preliminary experiments with extrapolation cases, the external calibration model's performance is fundamentally tethered to the accuracy of its simplistic linear shock predictor. When faced with unseen conditions outside the training distribution, this predictor fails, providing erroneous physical features to the neural network and causing a catastrophic failure in the prediction. In contrast, the end-to-end models learn a more fundamental, implicit relationship between the boundary conditions (PR) and the entire flow field. While slightly less accurate for interpolation, this learned representation is inherently more robust and possesses a greater potential for generalization. Therefore, the ablation study highlights a crucial trade-off: the external calibration method offers superior precision for interpolation, while the end-to-end approach provides the robustness and generalizability essential for a truly predictive surrogate model. Based on this analysis, the end-to-end model without relative weighting (Case 3) stands out as the most balanced and promising architecture for broader applications.

\subsection{Neural Model Validation}
We evaluate four neural operators for 2D nozzle flow prediction across back pressure-ratio conditions using a leave-one-out (LOO) protocol. All models take the PR as the branch input. Trunk inputs differ as follows:
\begin{itemize}
  \item \vanilla: classical DeepONet with a two-dimensional trunk $(x,y)$.
  \item \fusionorig: a multiplicative fusion of branch and trunk $(x,y)$ features using an MLP decoder.
  \item \currentmodel: a shock-aware fusion DeepONet that augments the trunk with nine physically--motivated features, including the signed distance to the predicted shock location, a smooth shock indicator, and multi-scale RBFs centered at the shock. Gaussian noise regularization and light weight decay are used.
  \item \unet: a 2D U-Net trained on zone grids; inputs are per-zone tensors constructed from the same nine features plus PR.
\end{itemize}
Training/validation splits preserve PR groups to avoid leakage. At test time, we evaluate on withheld PR cases. For Tecplot files, headers are preserved and only the $U,V$ fields are replaced; reported errors are computed pointwise from the ground truth fields present in the same file.

We report global relative $\ell_2$ errors for the streamwise ($U$), the cross-stream ($V$), and their joint vector $(U,V)$:
\begin{equation*}
\text{relL2}(a,\hat a) \;=\; \frac{\lVert a-\hat a\rVert_2}{\lVert a\rVert_2}\,.
\end{equation*}

\begin{table}[h!]
  \centering
  \caption{Global relative $\ell_2$ errors (lower is better).}
  \label{tab:metrics}
  \begin{tabular}{l *{3}{S[table-format=1.6]}}
    \toprule
    {Model} & {relL2\_U} & {relL2\_V} & {relL2\_joint} \\
    \midrule
    \vanilla     & 0.172439 & 0.092887 & 0.164621 \\
    \fusionorig  & 0.152038 & 0.080290 & 0.145033 \\
    \currentmodel\ & \bfseries 0.103905 & \bfseries 0.075993 & \bfseries 0.100844 \\
    \unet        & 0.503485 & 0.835374 & 0.555853 \\
    \bottomrule
  \end{tabular}
\end{table}

\noindent The \currentmodel\ wins because of the following reasons:

\noindent(1) \textbf{Shock localization in the trunk.} Augmenting $(x,y)$ with the signed distance to the shock, a smooth indicator, and multi-scale RBFs gives the trunk an explicit \emph{coordinate system aligned with the discontinuity}. This reduces the burden on the decoder to discover non-stationary features and improves interpolation across PR.\\
(2) \textbf{Multiplicative fusion.} Hadamard fusion between branch and trunk pathways lets the network \emph{gate} spatial responses by PR, which is crucial as shock position and strength vary with PR.\\
(3) \textbf{Stability near steep gradients.} Mild Gaussian input noise, Huber loss, and gradient-aware sample weighting attenuate the influence of outliers and reduce overfitting around the shock, improving generalization for both $U$ and $V$.

The baseline models underperform because:
\vanilla\ and \fusionorig\ operate in raw $(x,y)$, which is a poor coordinate system for shocks: the mapping from $(x,y, \text{PR})$ to $U,V$ is highly non-stationary and exhibits kinked responses near the discontinuity. Without shock-aligned features the models must learn a large set of location-specific filters, leading to higher global error---especially in $U$, which carries the dominant jump.\\[2pt]
\unet\ is further disadvantaged because (i) Tecplot zones do not form a single uniform image grid, so regridding introduces interpolation artifacts; (ii) convolution with fixed, translation-equivariant kernels is not well-suited to PR-dependent shock translations unless equipped with a strong \emph{positional/shock prior}; and (iii) skip connections can pass through local noise at the discontinuity, which raises the relative error for $V$ and the joint metric.

The remaining error for \currentmodel is concentrated in a thin band across the shock and at the outlet boundary. These are regions of large curvature and mild mesh anisotropy; the relative error denominator is also small there, which slightly inflates normalized metrics. Nevertheless, \currentmodel cuts the joint error by \textbf{${\sim}39\%$} versus \vanilla\ and by \textbf{${\sim}30\%$} versus \fusionorig.

\subsection{Nozzle with Throat Location Change}
\begin{figure}[t]
  \centering
  \includegraphics[width=0.95\linewidth]{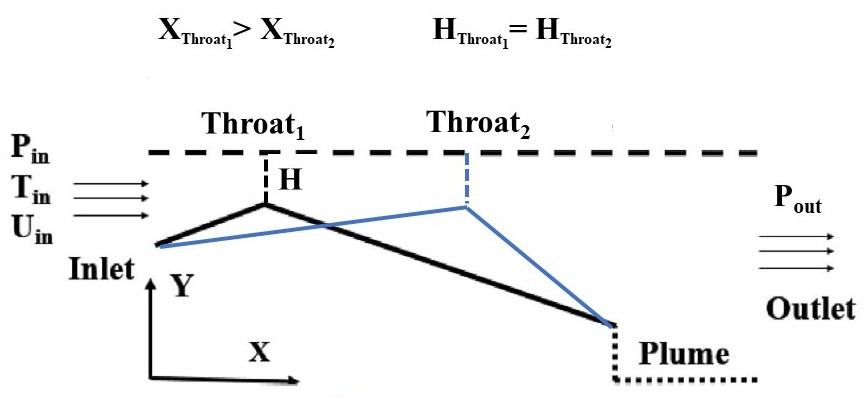}%
  \caption{Schematic of the micro-nozzle with downstream plume region with throat location change.}
  \label{fig:nozzle-schematic2}
\end{figure}

To quantify the sensitivity of the internal flow and plume to throat geometry, we vary only the axial throat position while keeping the throat height fixed, see Fig.~\ref{fig:nozzle-schematic2}. Specifically, the non-dimensional location is swept as
\(X_{\text{throat}}/L \in [0.10,\,0.55]\) at eight settings
(0.10, 0.15, 0.20, 0.25, 0.30, 0.35, 0.45, 0.55),
where \(L\) is the nozzle length. Inlet conditions \((P_{\text{in}}, T_{\text{in}}, U_{\text{in}})\) and outlet pressure \(P_{\text{out}}\) are identical for all cases. One configuration (e.g., \(X_{\text{throat}}/L=0.30\)) is held out for testing; the rest are used for training/validation.

\begin{figure}[t]
  \centering
  \begin{subfigure}[t]{0.49\textwidth}
    \centering
    \includegraphics[width=\linewidth]{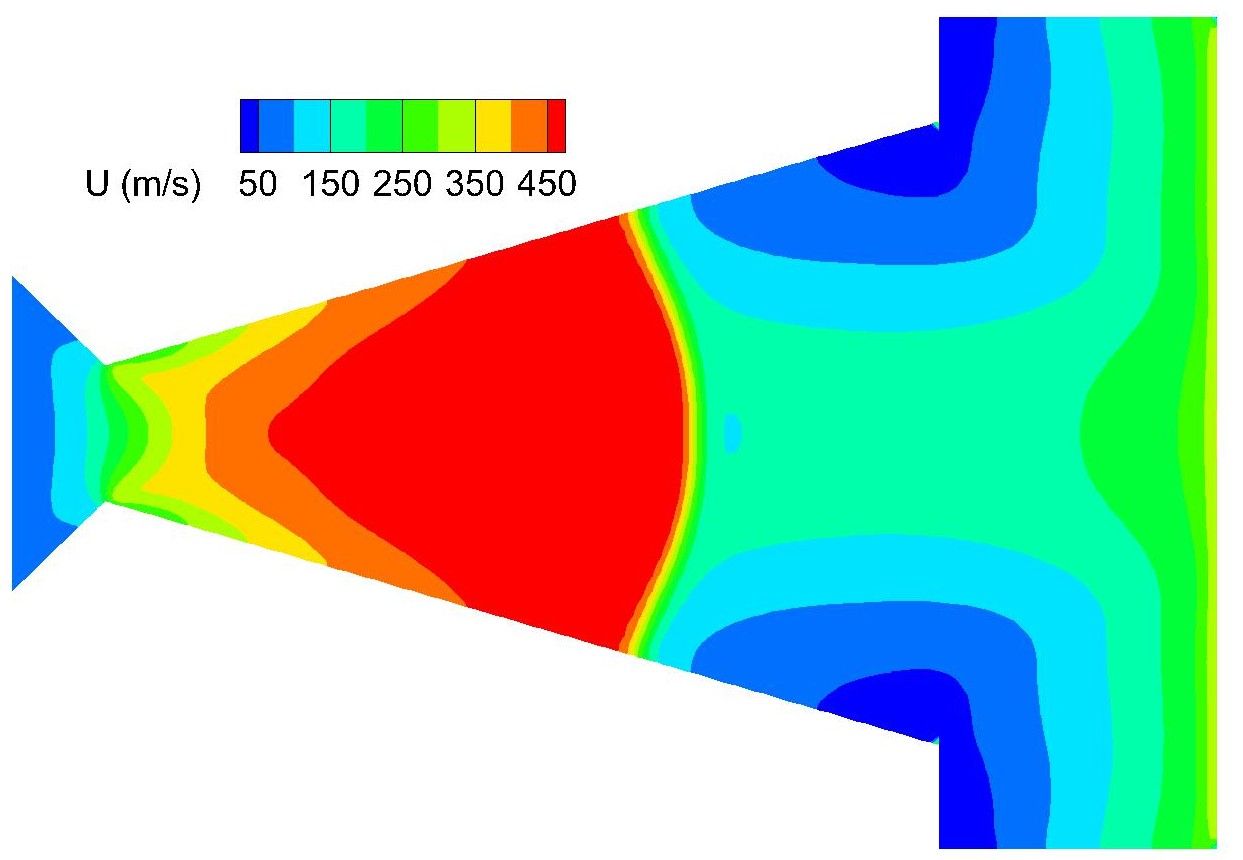}
    \caption{Throat at \(X_{\text{throat}}/L = 0.10\).}
    \label{fig:throat10}
  \end{subfigure}
  \hfill
  \begin{subfigure}[t]{0.49\textwidth}
    \centering
    \includegraphics[width=\linewidth]{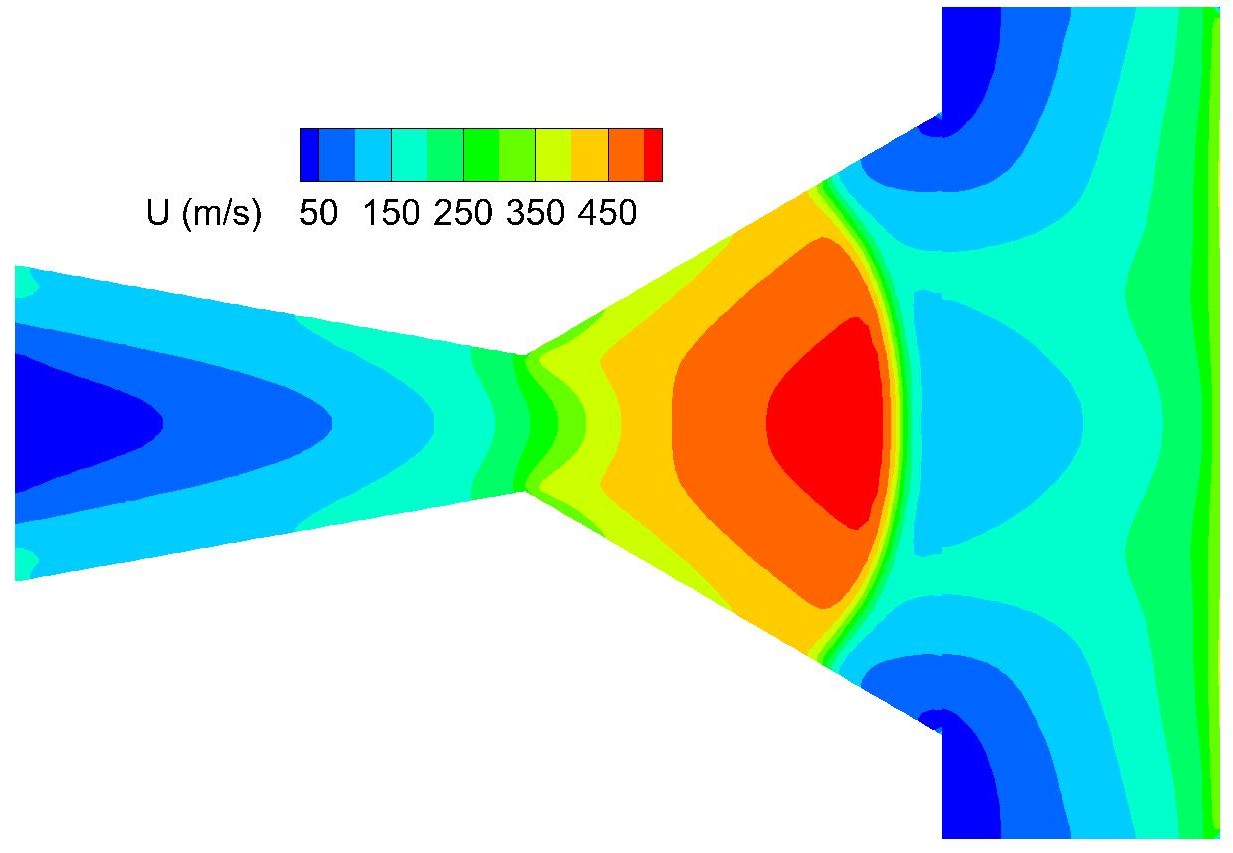}
    \caption{Throat at \(X_{\text{throat}}/L = 0.55\).}
    \label{fig:throat55}
  \end{subfigure}
  \caption{DSMC velocity–magnitude contours \(U~(\mathrm{m/s})\) for two extreme throat locations. All inlet and outlet conditions are identical; only the axial throat position is varied.}
  \label{fig:throat_sweep}
\end{figure}

To illustrate the sensitivity of the internal expansion and the termination shock to the throat position, Fig.~\ref{fig:throat_sweep} shows DSMC solutions at the two extremes of the sweep: \(X_{\text{throat}}/L=0.10\) and \(0.55\). With identical inlet/outlet conditions, moving the throat downstream (from 0.10 to 0.55) enlarges the high–velocity core in the divergent section and pushes the primary shock system farther downstream toward the exit/plume region. Conversely, with the throat near the inlet, the acceleration zone is shorter and the shock forms closer to the geometric expansion. These two cases bound the observed range of shock locations over the entire throat-position interval considered.

We use a lightweight operator-learning network with a branch--trunk fusion. 
The \emph{branch} takes a single scalar parameter, the non-dimensional throat location $X_{\text{throat}}/L$. 
The \emph{trunk} ingests local geometric/physical features at each grid point:
$(x,y)$, the axial offset to the estimated throat $d=x-x_t$, a logistic switch $s=\bigl(1+e^{-\kappa(x_t-x)}\bigr)^{-1}$, $|d|$, $d^2$, three Gaussian RBFs in $d$ (with widths $3\Delta x$, $7\Delta x$, $12\Delta x$), and the normalized distance to the nearest wall. 
The throat abscissa $x_t$ is calibrated from training cases via the $x$-location of the peak median $|\partial U/\partial x|$ at each $X_{\text{throat}}/L$ and then fitted with a linear map $x_t=a+b\,(X_{\text{throat}}/L)$.

Both branches are passed through three \texttt{Dense}+\texttt{LayerNorm}+\texttt{Swish}+\texttt{Dropout} blocks and fused by elementwise multiplication; a two–layer decoder maps the fused representation to $(U,V)$.

We train on all throat locations except the test case ($0.10$--$0.55$ step $0.05$, one held out). 
Mini-batches are drawn from all training fields; points near the throat and near the inlet are lightly oversampled with small $x$-jitter. 
Sample weights combine (i) a Gaussian distance-to-throat factor, (ii) a gradient-based factor from normalized $|\partial U/\partial x|$ to emphasize shocks, (iii) an inlet emphasis window, and (iv) a mild relative weighting to keep low-speed regions from being underweighted. 
Inputs/outputs are standard-scaled.

The Hyperparameters are as follows:
Hidden sizes: branch/trunk width \(=96\), decoder width \(=128\) (two layers). 
Regularization: L2 \(=3\times 10^{-4}\); dropout \(=0.35\); input Gaussian noise \(=0.03\). 
Optimizer: AdamW with initial LR \(8\times10^{-4}\) (clipnorm \(=1.0\)); batch size \(=512\). 
Loss: Huber with \(\delta=0.6\) for both warm-up and focus phases (to avoid loss-scale jumps), followed by a short fine-tune with a cosine-decayed learning rate. 
Learning-rate schedule: ReduceLROnPlateau in phase~1; cosine decay (no restarts) in phase~2. 
Validation split is group-wise (\(20\%\)) by throat configuration.

For each geometry, we use the DSMC fields on the same 2D structured grid: coordinates $(x,y)$ and velocity targets $(U,V)$. 
All inlet thermodynamic/kinematic conditions and the outlet pressure are identical across the sweep; only the axial throat position varies while the throat height is fixed. 
One configuration (here $X_{\text{throat}}/L=0.30$) is entirely held out for testing.

In the earlier model parameterized by pressure ratio $P_{\text{out}}/P_{\text{in}}$, the branch input was a \emph{flow condition}; the geometry was fixed and the trunk used $(x,y)$ and generic geometric features. 
Here the branch input is a \emph{geometric} parameter ($X_{\text{throat}}/L$); the trunk is augmented with throat-relative features ($d$, $s$, RBFs) and wall distance so the network is explicitly aware of where the throat—and hence the shock system—is expected. 
The weighting strategy is also adapted: we emphasize the throat neighborhood and the inlet (to reduce inlet bias) and use a gradient-informed weight tied to $U$’s axial derivative to better resolve shocks that migrate with throat position.

\begin{figure}[t]
  \centering
  \begin{subfigure}[t]{0.49\textwidth}
    \centering
    \includegraphics[width=\linewidth]{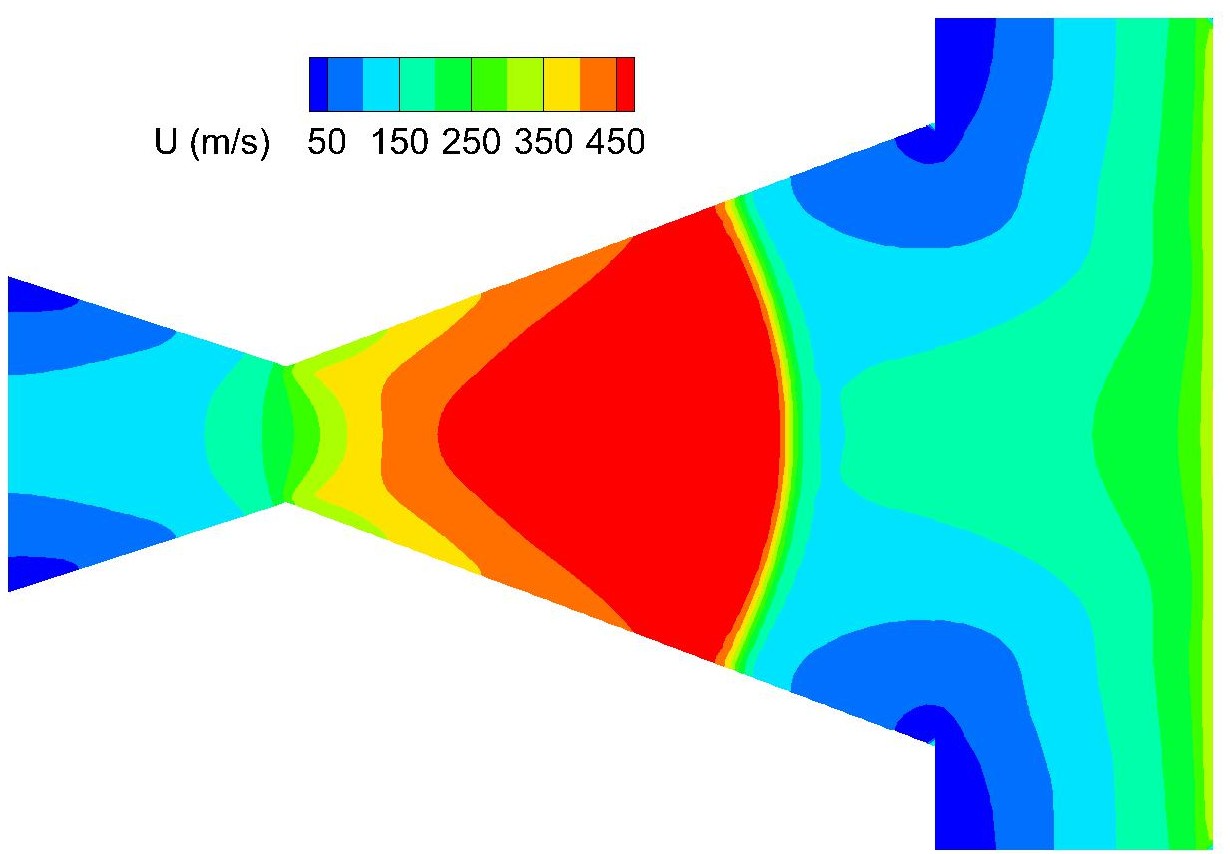}
    \caption{DSMC, $U$ (m/s) at $X_{\text{throat}}/L=0.30$.}
    \label{fig:U_dsmc_30}
  \end{subfigure}\hfill
  \begin{subfigure}[t]{0.49\textwidth}
    \centering
    \includegraphics[width=\linewidth]{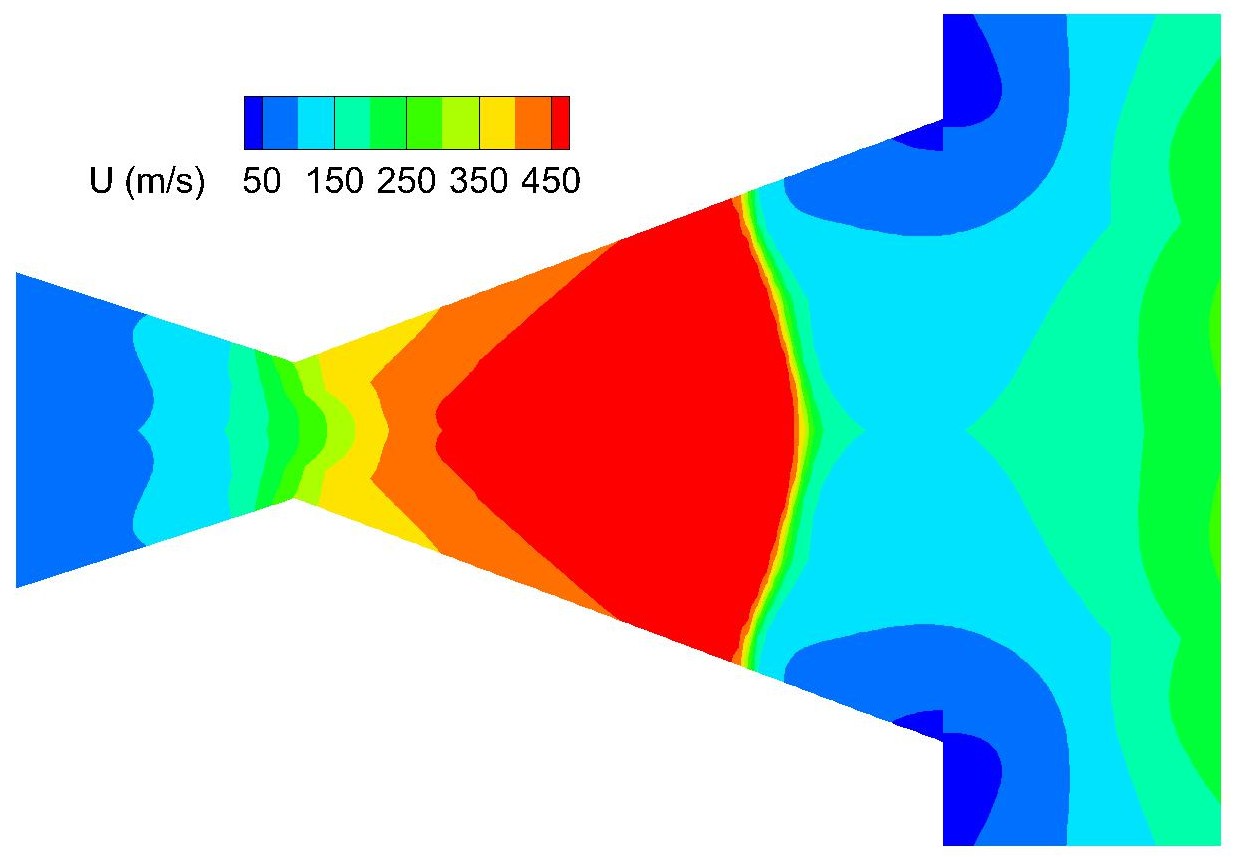}
    \caption{Neural network, $U$ (m/s) at $X_{\text{throat}}/L=0.30$.}
    \label{fig:U_nn_30}
  \end{subfigure}
  \caption{Velocity–magnitude contours for the streamwise component \(U\) at the test configuration \(X_{\text{throat}}/L=0.30\).}
  \label{fig:U_compare_30}
\end{figure}

\begin{figure}[t]
  \centering
  \begin{subfigure}[t]{0.49\textwidth}
    \centering
    \includegraphics[width=\linewidth]{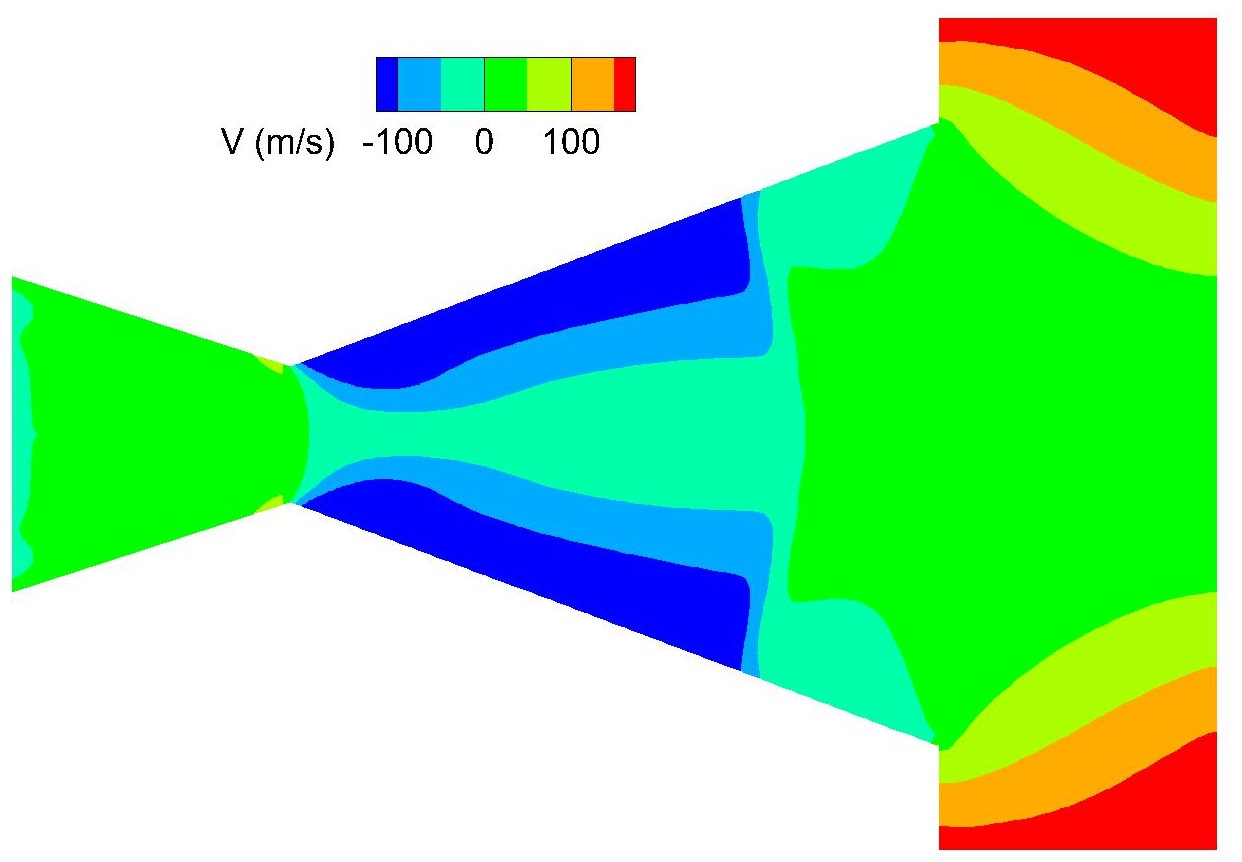}
    \caption{DSMC, $V$ (m/s) at $X_{\text{throat}}/L=0.30$.}
    \label{fig:V_dsmc_30}
  \end{subfigure}\hfill
  \begin{subfigure}[t]{0.49\textwidth}
    \centering
    \includegraphics[width=\linewidth]{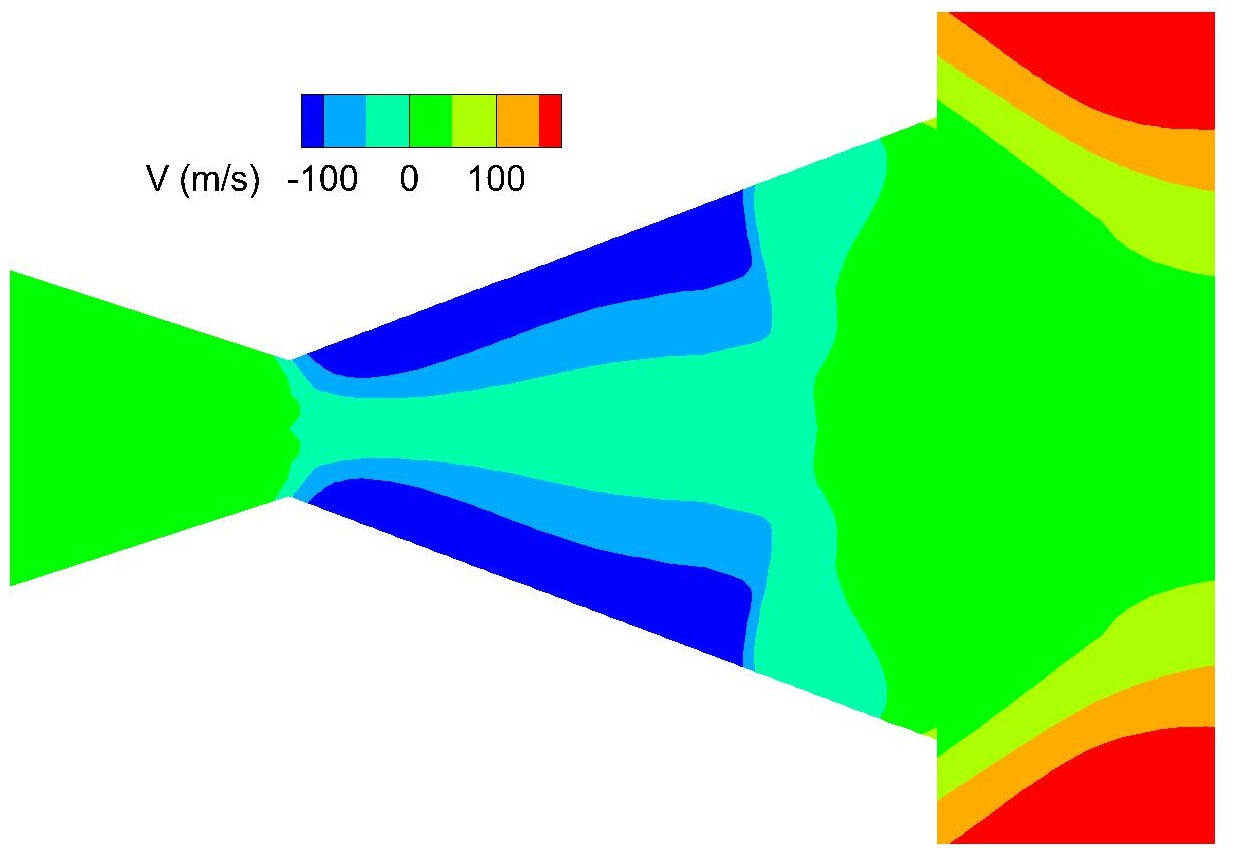}
    \caption{Neural network, $V$ (m/s) at $X_{\text{throat}}/L=0.30$.}
    \label{fig:V_nn_30}
  \end{subfigure}
  \caption{Cross–stream velocity \(V\) at the same test case. The network reproduces the sign change across the shear layers and the amplitude decay downstream.}
  \label{fig:V_compare_30}
\end{figure}

Figures~\ref{fig:U_compare_30}--\ref{fig:V_compare_30} compare DSMC reference solutions and the neural network predictions at the held-out configuration with the throat located at $X_{\text{throat}}/L=0.30$. For the streamwise velocity $U$, the network captures the expansion through the throat, the build-up of the high–velocity core in the divergent section, and the steep gradients across the internal shock. The downstream decay and the transverse asymmetry induced by the wall kink are also reproduced. For the cross–stream component $V$, the network correctly predicts the alternating positive/negative lobes associated with shear layers and shock curvature, together with the attenuation of $|V|$ toward the exit.

\begin{figure}[t]
  \centering
  \begin{subfigure}[t]{0.49\textwidth}
    \centering
    \includegraphics[width=\linewidth]{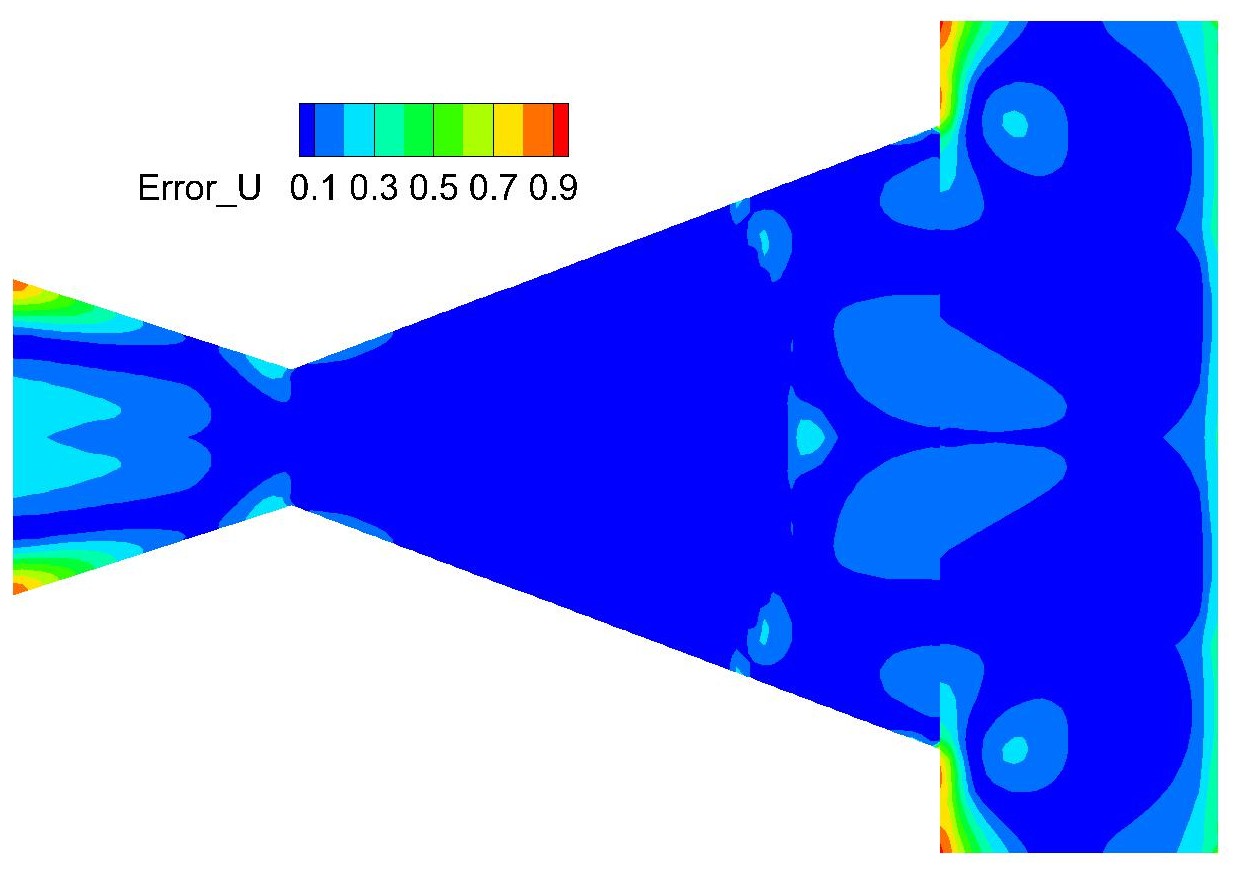} 
    \caption{Normalized error of $U$.}
    \label{fig:errU_30}
  \end{subfigure}
  \hfill
  \begin{subfigure}[t]{0.49\textwidth}
    \centering
    \includegraphics[width=\linewidth]{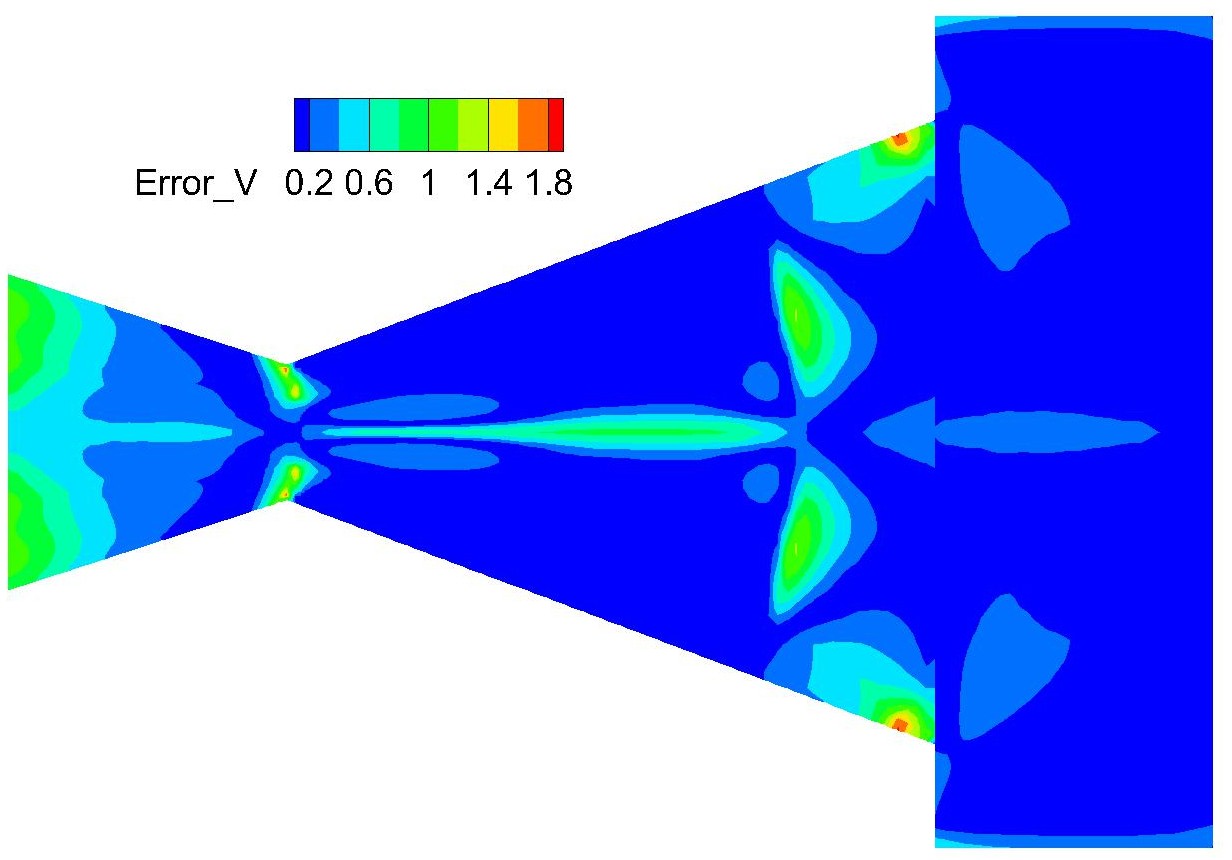} 
    \caption{Normalized error of $V$.}
    \label{fig:errV_30}
  \end{subfigure}
  \caption{Side-by-side error maps for the test case with $X_{\text{throat}}/L=0.30$. 
  Bright colors indicate higher normalized error; the largest values remain localized near the shock layer and the outlet corner.}
  \label{fig:errUVt_30}
\end{figure}

The normalized error map in Fig.~\ref{fig:errUVt_30} shows that the largest discrepancies are confined to thin regions with very high gradients: the shock layer and the outlet corner recirculation. Away from these features, the error remains small across most of the nozzle interior, which is consistent with the smoothness of the predicted contours.


\section{Concluding Remarks}
\label{sec:conclusion}

This study presented a comprehensive physics-guided operator-learning framework, termed the shock-aware Fusion--DeepONet, for constructing fast and accurate surrogates of rarefied micro-nozzle flows. The framework builds upon the Fusion--DeepONet architecture and enhances it with two synergistic innovations: a physics-guided trunk input embedding that encodes the flow field in a shock-aligned coordinate system, and a two-phase curriculum-learning strategy that dynamically emphasizes high-gradient regions. Together, these components enable the network to learn discontinuous and multi-regime flow behaviors with both stability and physical consistency—capabilities that conventional data-driven networks and continuum-based solvers struggle to achieve.

Quantitative and qualitative evaluations across multiple back pressures and nozzle geometries confirmed the strong predictive capability of the proposed surrogate. When trained for a single output parameter, the network reproduced DSMC reference fields with remarkable precision, capturing the internal shock location, curvature, and post-shock recovery with low epistemic uncertainty. When extended to multi-parameter learning, it successfully modeled velocity, Mach number, pressure, and temperature simultaneously, maintaining consistent physical trends despite a slight reduction in local accuracy. The extrapolation experiments at $P_{\mathrm{back}} = 33~\mathrm{kPa}$, beyond the range of training data, further demonstrated the model’s robustness: although small deviations appeared near the compression region, the network continued to predict shock displacement and overall flow topology correctly. This performance underscores its ability to generalize beyond interpolation regimes, a key requirement for predictive engineering surrogates.

The ablation study offered a detailed examination of each architectural and methodological component. It revealed that the multiplicative fusion between branch and trunk streams, the physically motivated trunk features, and the moderate network capacity were all essential for accurate shock reconstruction. Simplifying the architecture or removing the fusion layer led to significant performance degradation, confirming the necessity of multi-scale conditioning. Likewise, while explicit gradient-based weighting provided limited benefit for interpolation tasks, it enhanced robustness in extrapolative cases by guiding the network’s attention toward steep gradients. The comparison with the externally calibrated model clarified a fundamental trade-off: explicit feature engineering around a precomputed shock position yields higher precision in interpolation, whereas the end-to-end framework provides better generalization to unseen operating conditions. This insight will be valuable for future surrogate design across other rarefied flow systems.

The validation on nozzle geometries with variable throat location further highlighted the adaptability of the operator-learning approach. By reinterpreting the branch input as a geometric parameter and embedding throat-relative features into the trunk, the same neural operator could learn the complex coupling between geometry and flow behavior without retraining from scratch. The network accurately predicted the shift of the termination shock and the redistribution of velocity gradients across a broad range of throat positions, confirming its versatility for both parametric and geometric variations.

In summary, the shock-aware Fusion--DeepONet framework achieves three major goals: (i) it bridges the gap between high-fidelity kinetic solvers and practical design workflows by delivering orders-of-magnitude computational speed-up; (ii) it retains the physical fidelity required to resolve localized non-equilibrium structures such as shocks and shear layers; and (iii) it provides a scalable operator-learning formulation applicable to other rarefied or multi-regime systems. The combination of physics-based feature design, curriculum-driven optimization, and uncertainty quantification establishes a blueprint for building next-generation neural surrogates that are both interpretable and predictive.

From a computational perspective, the proposed framework provides a significant reduction in runtime compared with the baseline DSMC solver. A typical DSMC simulation of the micro-nozzle requires between 10 and 15 hours of CPU time, depending on the Knudsen number and boundary conditions. In contrast, training and inference of the proposed neural operator on high-end GPUs take less than 30 minutes in total for the same set of flow configurations. Although the hardware architectures differ (CPU for DSMC and GPU for the neural network), the comparison clearly illustrates the substantial computational efficiency and scalability benefits of the operator-learning framework, particularly for parametric studies or real-time flow control applications.

Future research will extend the present framework along several directions. First, the incorporation of thermochemical nonequilibrium and multi-species effects will allow direct modeling of plasma and chemically reactive micro-thrusters. Second, expanding the architecture to three-dimensional geometries will enable direct comparison with experimental plume measurements and facilitate integration into full spacecraft micro-propulsion design loops. Third, coupling the neural operator with adaptive sampling or active-learning strategies will reduce DSMC data requirements and automate dataset enrichment in regions of high epistemic uncertainty. Finally, embedding this surrogate within an uncertainty-aware optimization pipeline will open a pathway toward real-time design and control of rarefied propulsion devices.

Overall, the present work demonstrates that combining physics-guided inductive biases with operator-learning architectures offers a powerful and generalizable paradigm for modeling complex, discontinuous, and parameter-sensitive rarefied gas flows. The proposed approach not only accelerates simulation workflows but also deepens our understanding of how learning machines can internalize and reproduce the governing physics of shock-dominated systems.

\section*{Data Availability Statement}

The datasets and source codes that support the findings of this study for the Burger test case are publicly available at the following GitHub repository: 

\noindent\url{https://github.com/Ehsan-Roohi/Burger_Shock_Aware_Fusion_Deep_ONet}

This repository includes all Python scripts used for the Burgers equation solution. 

The DSMC simulation data and trained neural-operator models that support the findings of this study are available from the corresponding author upon reasonable request. Processed datasets and visualization scripts can be shared for research use.

\clearpage 
\bibliographystyle{unsrt}     
\bibliography{references}  

\end{document}